\documentclass{article}
\usepackage[OT1]{fontenc}

\usepackage{microtype}
\usepackage{graphicx}
\usepackage{subcaption}
\usepackage{booktabs}
\usepackage{amsmath,amssymb,tabularx,array}
\usepackage{hyperref}

\usepackage[preprint]{icml2026}

\usepackage{mathtools}
\usepackage{amsthm}
\usepackage[capitalize,noabbrev]{cleveref}

\theoremstyle{plain}
\newtheorem{theorem}{Theorem}[section]
\newtheorem{proposition}[theorem]{Proposition}

\theoremstyle{definition}

\theoremstyle{remark}

\icmltitlerunning{
Foresight: Learning Twisted SMC for Rare Events from Reversed Relaxations
}
\icmlcorrespondingauthor{Chen Fang}{fangchen23@ioz.ac.cn}

\hypersetup{pdfsubject={Preprint}}
\renewcommand{\today}{October 31, 2026}

\begin{document}

\twocolumn[
  \icmltitle{
  Foresight: Learning Twisted SMC for Rare Events from Reversed Relaxations
  }

  \icmlsetsymbol{equal}{*}

  \begin{icmlauthorlist}
    \icmlauthor{Ziyi Fan}{equal,ioz}
    \icmlauthor{Yuxuan Qiu}{equal,utokyo}
    \icmlauthor{Haibo Wang}{westlake}
    \icmlauthor{Wentao Cui}{blsa}
    \icmlauthor{Chen Fang}{ioz}
  \end{icmlauthorlist}

  \icmlaffiliation{ioz}{Key Laboratory of Organ Regeneration and Reconstruction, Chinese Academy of Sciences, Beijing, China}
  \icmlaffiliation{utokyo}{Institute of Industrial Science, The University of Tokyo, Tokyo, Japan}
  \icmlaffiliation{westlake}{Westlake University, Hangzhou, China}
  \icmlaffiliation{blsa}{Beijing Life Science Academy, Beijing, China}

  \icmlkeywords{
    Rare-event estimation,
    importance sampling,
    sequential Monte Carlo,
    reversible Markov chains
  }

  \vskip 0.3in
]

\printAffiliationsAndNotice{\icmlEqualContribution}
\raggedbottom

\begin{abstract}
We consider the probability that a reversible Markov chain started in
a set $A$ is in a rare set $B$ after a fixed horizon, a quantity often
needed for many start laws, down to single states.
Twisted sequential Monte Carlo (SMC) estimates such probabilities
without bias, with small variance when its proposal and twist
approximate the dynamics conditioned on reaching $B$, but these
components are usually learned from forward simulations, which
contain almost no successful paths when the event is rare.
We introduce Foresight-SMC, which learns both components from the
target side: under reversibility, relaxations started in $B$ that
return to $A$ are, read backwards, exact samples of the conditioned
paths, and their number is governed by the return probability rather
than by the rare probability.
Extending this reversal to the recorded propose--accept decisions of a
Metropolis-adjusted sampler, we fit a controlled Langevin proposal with
exact likelihood ratios by maximum likelihood, together with a
classifier twist; the resulting estimator is unbiased for any start
law.
A variance analysis shows that resampling reduces the variance when
the importance weights degenerate, as they do over long horizons.
On an Allen--Cahn field, the learned proposal alone suffices, and one
trained model is more accurate than splitting at equal cost.
On alanine tripeptide, resampling with the learned twist is necessary,
and Foresight-SMC needs less computation per query than adaptive
multilevel splitting on localized start sets.
\end{abstract}

\section{Introduction}
\label{sec:introduction}

Many questions about molecular and stochastic systems ask how likely a
rare transition is within a given time.
For a reversible Markov chain started from a law $\nu$ on a set $A$,
we seek the probability $p(\nu,T)$ that it is in a target set $B$
after $T$ steps.
Splitting methods \citep{allen2005sampling,cerou2007adaptive} and
importance sampling \citep{dupuis2004importance} estimate such
probabilities far below the reach of direct simulation.
Each run, however, answers a single start law and horizon, which we
call a query, and splitting further requires a reaction coordinate, a
scalar measure of progress toward $B$.
Applications often need many queries, down to single start
configurations.

Twisted sequential Monte Carlo (SMC)
\citep{delmoral2004feynman,chopin2020introduction} offers an
alternative that estimates such probabilities without bias and needs
no reaction coordinate.
It propagates particles under a proposal dynamics, corrects them with
likelihood ratios, and resamples them according to a twist, a function
that predicts how likely each particle is to reach $B$.
From a point start, the estimator has zero variance when the proposal
is the Doob $h$-transform, that is, the dynamics conditioned to end in
$B$, and the twist is the success probability $h$ itself
\citep{doob1957conditional,heng2020controlled}.
In practice both components are learned from forward simulations
\citep{guarniero2017iterated,heng2020controlled,lawson2022sixo},
which contain almost no successful paths when the event is rare.

We introduce Foresight-SMC (Figure~\ref{fig:method}), which learns the
proposal and the twist from the target side.
Our key observation is that reversibility supplies the missing
successful paths.
Consider relaxations, that is, reference paths started at equilibrium
in $B$. Those that are in $A$ after $T$ steps are, read backwards,
exact samples of the conditioned path law started from
$\mu_A=\pi(\cdot\mid A)$, the equilibrium law $\pi$ restricted to $A$.
Their fraction $r_B$ satisfies $\pi(A)\,p(\mu_A,T)=\pi(B)\,r_B$.
The rarity is therefore carried by the equilibrium ratio
$\pi(B)/\pi(A)$, whereas $r_B$ is not small once $T$ exceeds the
typical residence time in $B$.
The identity alone yields only the average over $\mu_A$ and requires
$\pi(B)/\pi(A)$. Foresight-SMC instead uses the returning relaxations
as training data, and the learned model answers queries from any start
law, including single configurations, without retraining.

Our contributions are as follows.
\begin{itemize}
\item We show that reversed relaxations are training data for twisted
SMC. We extend the classical time-reversal identity to the recorded
propose--accept decisions of a Metropolis-adjusted sampler, and show
that the effective sample size of the relaxations is at least their
number times $r_B$ (Section~\ref{sec:relaxation}).
\item We learn $h$ in the two forms that twisted SMC needs: a controlled
Langevin proposal whose recorded steps give exact likelihood ratios,
fitted by maximum likelihood on reversed transitions, and a classifier
twist whose logit is $\log h$ up to a constant. The estimator is
unbiased for any start law, including point starts, whatever the
networks learn (Sections~\ref{sec:controlled_kernel}--\ref{sec:smc}).
\item We relate the variance to the errors of the proposal and the
twist. The proposal error accumulates over the horizon, and resampling
with the twist resets the weights at the price of the twist error
(Section~\ref{sec:horizon}).
\end{itemize}
We evaluate Foresight-SMC on the two sides of this analysis
(Section~\ref{sec:experiments}). On an Allen--Cahn field, the learned
proposal alone suffices, and one trained model answers point queries
more accurately than splitting at equal cost. On alanine tripeptide,
the weights of the proposal alone degenerate over long paths, and
resampling with the learned twist needs less work per query than
splitting on localized start sets.

\section{Preliminaries}
\label{sec:preliminaries}

\paragraph{Problem setup.}
We consider a discrete-time Markov chain on a state space $\mathcal{X}$
with transition kernel $K$, which we call the reference dynamics;
$K(x,dy)$ is the probability of moving from $x$ into $dy$ in one step.
For a function $f$ we write $Kf(x)=\int K(x,dy)\,f(y)$, so that
$K^s f(x)$ is the expectation of $f(X_s)$ for the chain started at
$x$; for a measure $\mu$, we write $\mu(f)=\int f\,d\mu$, and
$\mu K^s$ denotes the law of $X_s$ when $X_0\sim\mu$.
We assume that $K$ is reversible with respect to a probability
measure $\pi$,
\begin{equation}
\label{eq:reversibility}
\pi(dx)\,K(x,dy)=\pi(dy)\,K(y,dx),
\end{equation}
which implies that $\pi$ is stationary for $K$.
Let $A,B\subset\mathcal{X}$ be disjoint start and target sets with
$\pi(A),\pi(B)>0$, and write $\mu_A=\pi(\cdot\mid A)$ and
$\mu_B=\pi(\cdot\mid B)$.

A query consists of a start law $\nu$ on $A$, either a point mass
$\delta_{x_0}$ (a point start) or a distribution over start states,
and a horizon $T\in\mathbb{N}$.
Let $P_\nu$ denote the law of the path $X_{0:T}=(X_0,\dots,X_T)$
with $X_0\sim\nu$ and $X_{t+1}\sim K(X_t,\cdot)$.
We seek the endpoint probability
\begin{equation}
\label{eq:query_probability}
\begin{aligned}
p(\nu,T) &= P_\nu(X_T\in B)=\int h_T(x)\,\nu(dx),\\
h_s(x)   &= K^s\mathbf{1}_B(x)=P_{\delta_x}(X_s\in B),
\end{aligned}
\end{equation}
where $\mathbf{1}_B$ denotes the indicator function of $B$, so that
$h_s(x)$ is the probability of being in $B$ after $s$ further steps
from $x$.
Note that the event is $X_T\in B$ rather than the first arrival in
$B$ before time $T$; the reversal identity of
Section~\ref{sec:relaxation} holds for this endpoint event.

\paragraph{Reference dynamics.}
In our experiments, $\pi(x)\propto e^{-U(x)}$ for a potential energy
$U$ on $\mathcal{X}=\mathbb{R}^d$ or on a torus, and $K$ is the
Metropolis-adjusted Langevin algorithm
\citep[MALA;][]{roberts1996exponential}.
From a state $x$, MALA proposes an Euler--Maruyama step of the
overdamped Langevin diffusion $dX_t=-\nabla U(X_t)\,dt+\sqrt{2}\,dW_t$,
where $W_t$ is a standard Brownian motion. With step size $\tau>0$,
the proposal density is
\begin{equation}
\label{eq:mala_proposal}
q(x,z)=\mathcal{N}\big(z;\,x-\tau\nabla U(x),\,2\tau I\big),
\end{equation}
where $\mathcal{N}(z;m,\Sigma)$ is the Gaussian density in $z$ and
$I$ is the identity.
The proposal is accepted with the Metropolis--Hastings probability
\begin{equation}
\label{eq:mala_acceptance}
\alpha(x,z)=\min\left\{1,\;
\frac{e^{-U(z)}\,q(z,x)}{e^{-U(x)}\,q(x,z)}\right\},
\end{equation}
and the chain remains at $x$ otherwise.
The acceptance step makes $K$ exactly reversible with respect to
$\pi$ for every $\tau$, as required by \eqref{eq:reversibility}.
The rejection probability $1-\int q(x,z)\,\alpha(x,z)\,dz$, however,
has no closed form, so a MALA path has no tractable density;
Section~\ref{sec:relaxation} handles this by recording the
propose--accept decisions.

\paragraph{Conditioned path law.}
For $p(\nu,T)>0$, conditioning $P_\nu$ on the terminal event gives
\begin{equation}
\label{eq:conditioned_path_law}
P^\star_\nu = P_\nu(\,\cdot\mid X_T\in B),
\qquad
\frac{dP^\star_\nu}{dP_\nu}=\frac{\mathbf{1}_B(X_T)}{p(\nu,T)},
\end{equation}
that is, $P^\star_\nu$ keeps only the paths that end in $B$ and
renormalizes their probability.
By the Doob $h$-transform~\citep{doob1957conditional},
$P^\star_\nu$ is again a Markov chain, with initial law and
time-dependent kernels
\begin{equation}
\label{eq:doob_kernel}
\begin{aligned}
\nu^\star(dx)   &= \frac{h_T(x)}{p(\nu,T)}\,\nu(dx),\\
K_t^\star(x,dy) &= K(x,dy)\,\frac{h_{T-t-1}(y)}{h_{T-t}(x)},
\end{aligned}
\end{equation}
defined where $h_{T-t}(x)>0$.
Here $K_t^\star$ governs the step from time $t$ to $t+1$, when $T-t$
steps remain. It reweights each reference step by the success
probability $h_{T-t-1}(y)$ of the state it reaches; the denominator is
the normalizing constant, since $h_{T-t}=K h_{T-t-1}$.
In importance sampling, one draws paths from a proposal $Q$ and
averages the weights $\mathbf{1}_B(X_T)\,dP_\nu/dQ$, whose mean is
$p(\nu,T)$ when $P_\nu\ll Q$ on $\{X_T\in B\}$.
The conditioned law is the zero-variance proposal: with
$Q=P^\star_\nu$, every weight equals $p(\nu,T)$ exactly.
Its kernels, however, depend on the unknown functions $h_s$, about
which forward simulation is nearly uninformative when $p(\nu,T)$ is
small.

\section{Foresight-SMC}
\label{sec:method}

\begin{figure*}[t]
\centering
\includegraphics[width=\linewidth]{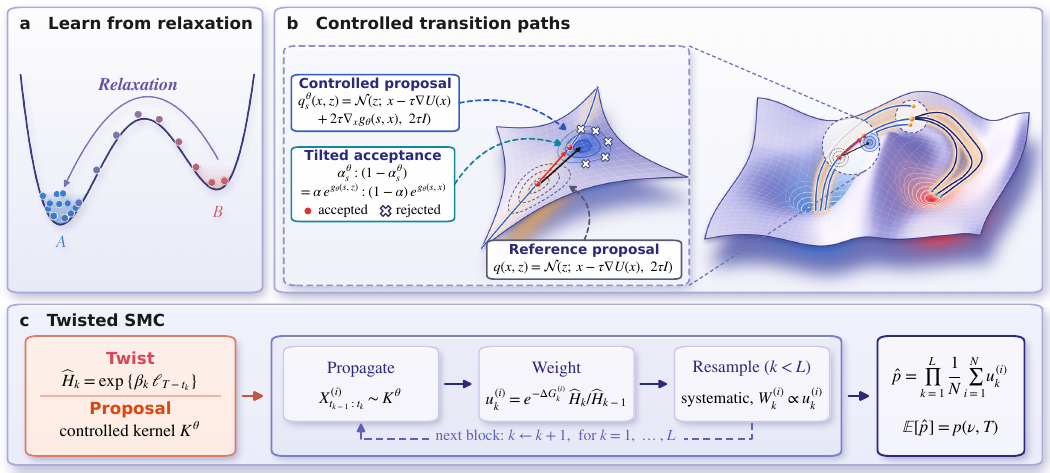}
\caption{Overview of Foresight-SMC.
(a)~Relaxations start from $Y_0\sim\mu_B$; those that end in $A$, read
backwards, are samples of $P^\star_{\mu_A}$
(Proposition~\ref{prop:path_reversal}) and train the proposal and the
twist.
(b)~For a query, particles move from $A$ (blue) toward $B$ (red) under
the controlled kernel $K^\theta$ and are resampled at block ends.
Inset: one step, in which the reference Gaussian proposal (dashed) is
shifted by $2\tau\nabla_x g_\theta$ and the acceptance is tilted by
$e^{g_\theta}$; crosses mark rejected proposals.
(c)~Twisted SMC with proposal $K^\theta$ and twist $\widehat H_k$
returns an unbiased estimate of $p(\nu,T)$; with $L=1$ it reduces to
importance sampling.}
\label{fig:method}
\end{figure*}

Section~\ref{sec:preliminaries} shows that the conditioned path law is
the zero-variance proposal but depends on the unknown success
probabilities $h_s$.
Foresight-SMC learns approximations of the Doob kernel and of $h_s$
from relaxations started in $B$ and combines them in twisted SMC
(Figure~\ref{fig:method}).
Section~\ref{sec:relaxation} obtains training paths by reversing
relaxations (Figure~\ref{fig:method}a).
Sections~\ref{sec:controlled_kernel} and~\ref{sec:twist} learn the
proposal (inset of Figure~\ref{fig:method}b) and the twist
(Figure~\ref{fig:method}c, left), Section~\ref{sec:smc} assembles the
estimator (Figure~\ref{fig:method}b,~c), and Section~\ref{sec:horizon}
analyzes its variance.
All proofs are given in the appendix.

\subsection{Reversed relaxations}
\label{sec:relaxation}

Both the proposal and the twist are learned from paths of the
conditioned law $P^\star_{\mu_A}$, and we obtain these paths from the
target side.
We draw $Y_0\sim\mu_B$ and run the reference chain for $T$ steps.
We call $Y_{0:T}$ a relaxation, $Y_s$ its state at age~$s$, and write
$\bar\mu_s=\mu_BK^s$; a relaxation is returning if $Y_T\in A$, with
return probability $r_B=P_{\mu_B}(Y_T\in A)$.
Sampling $\mu_B$ is an equilibrium problem restricted to $B$
(Appendices~\ref{app:ac} and~\ref{app:ala3}).
Read backwards, a returning relaxation is a path $X_t=Y_{T-t}$ from
$A$ to $B$ whose state at time $t$ has age $T-t$, the number of
remaining steps.

\paragraph{Transition records.}
As noted in Section~\ref{sec:preliminaries}, a MALA path has no
tractable density, yet twisted SMC needs likelihood ratios between the
learned and the reference kernel.
We therefore record each transition as $\xi=(z,a)$, with $z$ the
proposed state and $a\in\{0,1\}$ the acceptance indicator.
With respect to the product of $dz$ and the counting measure on
$\{0,1\}$, the record has the explicit density
\begin{equation}
\label{eq:record_density}
R(x,\xi)=q(x,z)\,\alpha(x,z)^{a}\,\big[1-\alpha(x,z)\big]^{1-a},
\end{equation}
so likelihood ratios between kernels with this propose--accept
structure are exact on recorded paths (Appendix~\ref{app:ala3_system} treats
the torus).
From now on, $P_\nu$ and $P^\star_\nu$ denote laws of recorded paths
$(X_0,\xi_{0:T-1})$.

The following proposition shows that reversal turns returning
relaxations into exact samples of the conditioned law, both for states
and for records.

\begin{proposition}[Path reversal]
\label{prop:path_reversal}
Under \eqref{eq:reversibility}, if $r_B>0$, then
\begin{equation}
\label{eq:path_reversal}
\operatorname{Law}\big(Y_T,\dots,Y_0 \,\big|\, Y_T\in A\big)=P^\star_{\mu_A},
\end{equation}
and $p(\mu_A,T)=\rho\,r_B$ with $\rho=\pi(B)/\pi(A)$.
If $K$ is MALA, \eqref{eq:path_reversal} also holds for recorded paths.
Reversal then keeps a rejected record unchanged and replaces the
proposed state of an accepted record by the state it left.
\end{proposition}

The statement for states and the identity are classical
\citep{kent1978time,e2006towards}: by \eqref{eq:reversibility},
equilibrium paths are invariant under time reversal, and restricting
the start to $B$ and the end to $A$ gives $P^\star_{\mu_A}$
(Appendix~\ref{app:path_reversal}).
The identity yields a baseline estimator for $\mu_A$, reverse counting
(Section~\ref{sec:exp_ala3}), but not $p(\nu,T)$ for other start laws.
The statement for records is new (Appendix~\ref{app:records}) and
allows the reversed records to serve as training data for the
proposal.

\paragraph{Coverage.}
It remains to quantify how informative these paths are.
The time-$t$ marginal $\nu^\star_t$ of $P^\star_{\mu_A}$ satisfies
$d\nu^\star_t/d\bar\mu_{T-t}\le 1/r_B$ (Appendix~\ref{app:coverage}),
so $n$ relaxation states at age $T-t$ have an effective sample size
of at least $n\,r_B$ for $\nu^\star_t$.
The data requirement is thus set by $r_B$ rather than by
$p(\mu_A,T)$.
The bound concerns $\mu_A$; for point starts, accuracy relies on the
generalization of the learned components.

\subsection{Learning the proposal}
\label{sec:controlled_kernel}

The Doob kernel \eqref{eq:doob_kernel} reweights a reference step at
time $t$ by the success probability $h_s(y)$ of the state $y$ that it
reaches, with $s=T-t-1$.
We approximate $\log h_s$, up to an additive constant, by a network
$g_\theta(s,\cdot)$, and reweight the proposal and the acceptance of a
MALA step separately (inset of Figure~\ref{fig:method}b).
Since $h_s$ does not depend on $T$, one network serves every horizon
up to the longest one used in training.

\paragraph{Proposal.}
Tilting $q(x,z)$ by $e^{g_\theta(s,z)}$ and linearizing $g_\theta$
around $x$ gives
\begin{equation}
\label{eq:controlled_proposal}
q^\theta_s(x,z)=
\mathcal{N}\big(z;\,x-\tau\nabla U(x)+2\tau\nabla_x g_\theta(s,x),\,2\tau I\big),
\end{equation}
whose added drift points toward states with a higher estimated
success probability.
In our experiments, the proposal is a mixture of two such Gaussians,
which lets paths follow distinct routes to $B$, the transition
channels (Appendix~\ref{app:implementation}).

\paragraph{Acceptance.}
Tilting the accept and reject outcomes by $e^{g_\theta(s,\cdot)}$ at
the resulting state gives, without linearization,
\begin{equation}
\label{eq:controlled_acceptance}
\alpha^\theta_s(x,z)=
\frac{\alpha(x,z)\,e^{g_\theta(s,z)}}
{\alpha(x,z)\,e^{g_\theta(s,z)}+\big(1-\alpha(x,z)\big)\,e^{g_\theta(s,x)}}.
\end{equation}
Substituting $q^\theta_s$ and $\alpha^\theta_s$ into
\eqref{eq:record_density} gives the record density $R^\theta_t$ of the
controlled kernel $K^\theta_t$.
We write $P^\theta_\nu$ for the law of its recorded paths from $\nu$,
and
\begin{equation}
\label{eq:log_likelihood_ratio}
G_\theta=\sum_{t=0}^{T-1}
\log\frac{R^\theta_t(X_t,\xi_t)}{R(X_t,\xi_t)}
\end{equation}
for their log-likelihood ratio to the reference dynamics.
We write $K^\theta$ for the family $(K^\theta_t)_{t<T}$.

\paragraph{Training.}
We minimize the forward KL divergence, which by
\eqref{eq:conditioned_path_law} is
\begin{equation}
\label{eq:kl_identity}
D_{\mathrm{KL}}\big(P^\star_{\mu_A}\,\big\|\,P^\theta_{\mu_A}\big)
=-\log p(\mu_A,T)-\mathbb{E}_{P^\star_{\mu_A}}\big[G_\theta\big].
\end{equation}
The first term does not depend on $\theta$, and by
Proposition~\ref{prop:path_reversal} the expectation can be estimated
on reversed records, which gives the cross-entropy loss
\citep{zhang2014applications,nusken2021solving}
\begin{equation}
\label{eq:loss}
\mathcal{L}(\theta)=-\,\mathbb{E}_{\mathrm{rev}}\big[G_\theta\big],
\end{equation}
where $\mathbb{E}_{\mathrm{rev}}$ averages over reversed records.
This loss is the negative log-likelihood of the reversed transitions
and needs no simulation under $K^\theta$ (Appendix~\ref{app:kl}).
Since $p(\mu_A,T)=\rho\,r_B$, evaluating \eqref{eq:kl_identity} on
held-out relaxations gives the path KL when $\rho$ is known, which
serves as a training diagnostic; for the reference dynamics,
$G_\theta\equiv0$ and it equals $-\log p(\mu_A,T)$.

\subsection{Learning the twist}
\label{sec:twist}

At a resampling time with $s$ steps remaining (Section~\ref{sec:smc}),
the twist needs to estimate $h_s$ only up to a positive factor, since
rescaling a twist at an intermediate resampling time leaves the
estimator unchanged (Appendix~\ref{app:classifier}).
One estimate is $e^{g_\theta(s,\cdot)}$.
Another is a classifier (Figure~\ref{fig:method}c, left): by
\eqref{eq:reversibility}, a relaxation state at age $s$ has density
$h_s/\pi(B)$ relative to $\pi$, so a balanced classifier between such
states and draws of $\pi$ has Bayes-optimal logit
\begin{equation}
\label{eq:classifier_logit}
\ell_s(x)=\log h_s(x)+c,\qquad c=-\log\pi(B)
\end{equation}
\citep[cf.][]{lawson2022sixo}.
With $\pi(B)$ known, $\pi(B)\,\nu(e^{\ell_T})$ is a simulation-free
plug-in estimate of $p(\nu,T)$, but it is biased under miscalibration
and comes without an error bar; using the classifier as a twist in
SMC removes both limitations.
Since any positive twist keeps the estimator unbiased, we may also
temper the classifier, $\exp\{\beta(s)\,\ell_s\}$ with $\beta(s)\ge1$,
to sharpen logits that are too small in magnitude
(Appendix~\ref{app:ala3_estimators}).
At the resampling time $t_k$ of Section~\ref{sec:smc}, we write
$\beta_k=\beta(T-t_k)$.
Algorithm~\ref{alg:train} summarizes training.

\begin{algorithm}[t]
\caption{Learning the components (once for all queries)}
\label{alg:train}
\begin{algorithmic}[1]
\REQUIRE reference kernel $K$; samplers of $\mu_B$ and $\pi$;
longest horizon $T$; pool size $n$
\FOR{$m=1$ \textbf{to} $n$}
\STATE Draw $Y_0\sim\mu_B$ and simulate $Y_{1:T}$ under $K$,
recording $\xi_{0:T-1}$
\ENDFOR
\STATE Reverse the paths with $Y_T\in A$, together with their records
(Proposition~\ref{prop:path_reversal})
\STATE Fit $g_\theta$ by minimizing \eqref{eq:loss} on the reversed
paths; this defines $K^\theta$
\STATE Fit $\ell_s$ by classifying states of age $s$ of all
relaxations against draws of $\pi$ (logit \eqref{eq:classifier_logit})
\ENSURE controlled kernel $K^\theta$ and logits $\ell_s$, from which each
query forms $\widehat H_k=\exp\{\beta_k\ell_{T-t_k}\}$ (or
$e^{g_\theta(T-t_k,\cdot)}$)
\end{algorithmic}
\end{algorithm}

\subsection{The twisted SMC estimator}
\label{sec:smc}

Twisted SMC propagates $N$ particles under $K^\theta$ and resamples
them at times $0=t_0<\dots<t_L=T$, which divide the horizon into $L$
blocks (Figure~\ref{fig:method}b).
At $t_k$, the twist $\widehat H_k$ is one of the estimates of
$H_k=h_{T-t_k}$ from Section~\ref{sec:twist}, and
$\widehat H_0\equiv1$, $\widehat H_L=\mathbf{1}_B$.
Each block consists of the three steps of Figure~\ref{fig:method}c.
Particles start from the states assigned by the previous resampling,
or from $\nu$, and are propagated under $K^\theta$ to $t_k$; particle
$i$ receives the weight
\begin{equation}
\label{eq:incremental_weight}
u_k^{(i)}=e^{-\Delta G_k^{(i)}}\,
\frac{\widehat H_k\big(X^{(i)}_{t_k}\big)}{\widehat H_{k-1}\big(X^{(i)}_{t_{k-1}}\big)},
\end{equation}
where $\Delta G_k^{(i)}$ is the increment of $G_\theta$ over the block.
If $k<L$, particle $i$ is then replaced by $O_k^{(i)}$ copies, with
offspring counts drawn according to the normalized weights
$W_k^{(i)}\propto u_k^{(i)}$.
The estimator is
\begin{equation}
\label{eq:smc_estimator}
\hat p=\prod_{k=1}^{L}\frac{1}{N}\sum_{i=1}^{N}u_k^{(i)}.
\end{equation}
The factor $e^{-\Delta G_k}$ corrects for the deviation from the
reference dynamics, and, at resampling, the twist ratio favors the
particles with a high estimated probability of reaching $B$.
Algorithm~\ref{alg:smc} gives the procedure for one query.

\begin{algorithm}[t]
\caption{Foresight-SMC for one query}
\label{alg:smc}
\begin{algorithmic}[1]
\REQUIRE query $(\nu,T)$; controlled kernel $K^\theta$; twists
$\widehat H_{0:L}$ with $\widehat H_0\equiv1$, $\widehat H_L=\mathbf{1}_B$;
block times $t_{0:L}$; $N$ particles
\STATE Draw $X^{(i)}_0\sim\nu$ for $i=1,\dots,N$; set $\hat p\gets1$
\FOR{$k=1$ \textbf{to} $L$}
\STATE Propagate each particle from $t_{k-1}$ to $t_k$ under
$K^\theta$, recording $\xi_t$, and accumulate $\Delta G^{(i)}_k$
\STATE $u^{(i)}_k\gets e^{-\Delta G^{(i)}_k}\,
\widehat H_k(X^{(i)}_{t_k})/\widehat H_{k-1}(X^{(i)}_{t_{k-1}})$
\STATE $\hat p\gets\hat p\cdot N^{-1}\sum_{i}u^{(i)}_k$
\IF{$k<L$}
\STATE Resample the particle histories with probabilities
$\propto u^{(i)}_k$ (systematic)
\ENDIF
\ENDFOR
\ENSURE unbiased estimate $\hat p$ of $p(\nu,T)$
\end{algorithmic}
\end{algorithm}

The estimator is unbiased however accurate the learned components are.

\begin{theorem}[Unbiasedness]
\label{thm:unbiased}
Assume $R^\theta_t(x,\cdot)>0$ wherever $R(x,\cdot)>0$, and let the
interior twists $\widehat H_k$, $0<k<L$, be measurable, finite,
positive and fixed independently of the particle system.
Let $\mathcal{F}_k$ denote the particle histories up to $t_k$, and
suppose the offspring counts satisfy
$\mathbb{E}[O_k^{(i)}\mid\mathcal{F}_k]=N\,W_k^{(i)}$ and the
resampling randomness is independent of later propagation.
Then $\mathbb{E}[\hat p]=p(\nu,T)$ for every $N$.
\end{theorem}

The result is standard for Feynman--Kac models
\citep{delmoral2004feynman}: the twist ratios telescope, so the
weights along a path multiply to $\mathbf{1}_B(X_T)\,e^{-G_\theta}$;
the proof in Appendix~\ref{app:unbiased} also covers the systematic
resampling that we use.
With $L=1$, \eqref{eq:smc_estimator} is importance sampling.
With the Doob kernel written on records
(Appendix~\ref{app:implementation}) and $\widehat H_k\propto H_k$, all
particles receive equal weights after the first block, and
$\hat p=p(\nu,T)$ for every $N$ from a point start.
The learned components only approximate this ideal, and
Section~\ref{sec:horizon} quantifies the cost of the approximation.

\subsection{Variance analysis}
\label{sec:horizon}

We relate the variance of $\hat p$ to the errors of the two
components.
With a single block,
\begin{equation}
\label{eq:is_variance}
\frac{N\operatorname{Var}(\hat p)}{p(\nu,T)^2}
=\chi^2\big(P^\star_\nu\,\|\,P^\theta_\nu\big)
\ge e^{D_{\mathrm{KL}}(P^\star_\nu\|P^\theta_\nu)}-1
\end{equation}
\citep{chatterjee2018sample,agapiou2017importance}, so the number of
particles needed for a fixed relative error grows at least
exponentially in the path KL.
By the chain rule, this KL is a sum of per-step divergences, which
accumulate over long horizons.
Resampling resets the weights at each boundary. For a point start
whose path KL $\kappa$ splits into $\kappa_k$ over the blocks, the
lower bound $e^{\kappa}-1$ for $L=1$ becomes
$\sum_k(e^{\kappa_k}-1)\ge L(e^{\kappa/L}-1)$, which grows only
linearly in $\kappa$ when $L\propto\kappa$. The second bound applies to
the asymptotic variance under multinomial resampling.
The price is an additional variance term due to the twist error, the
mismatch between $\widehat H_k$ and $H_k$ up to scale, which vanishes
only if $\widehat H_k\propto H_k$ on the states visited by conditioned
paths (Appendix~\ref{app:variance}).
We therefore expect resampling to help when the weights without
resampling degenerate, as on alanine tripeptide, and otherwise mainly
to add twist error, as on the Allen--Cahn field
(Section~\ref{sec:experiments} and Appendix~\ref{app:lsweep}).

\section{Experiments}
\label{sec:experiments}

We evaluate Foresight-SMC on two systems that correspond to the two
regimes of Section~\ref{sec:horizon}.
On a one-dimensional Allen--Cahn field (Section~\ref{sec:ac}), the
learned proposal alone is accurate.
On a model of alanine tripeptide over $16000$ steps
(Section~\ref{sec:exp_ala3}), it is not, and resampling becomes
necessary.
In both, one trained model answers all queries, and cost is the
number of evaluations of $U$ and $\nabla U$;
Table~\ref{tab:hparams} lists the settings.

\subsection{Allen--Cahn field}
\label{sec:ac}

The Allen--Cahn equation \citep{allen1979microscopic} models phase
separation: a scalar field relaxes toward one of two phases,
$u\approx+1$ or $u\approx-1$, separated by thin interfaces.
Under thermal noise, a droplet of one phase can nucleate inside the
other, a standard rare event for stochastic fields
\citep{kohn2007action}.
We discretize the field on a chain of $64$ sites, tilt the energy so
that the phase $u\approx+1$ is favored, and use MALA as the reference
dynamics (Appendix~\ref{app:ac}).
The rare event is the nucleation of a droplet of the unfavored phase:
with $\mathcal{D}$ the length of the longest run of negative sites,
$A=\{\mathcal{D}\le3\}$ and $B=\{10\le\mathcal{D}\le16\}$.
We query $30$ point starts in $A$ and three start laws ($\pi$
restricted to droplet sizes $0$, $1$ and $2$--$3$), each at $T=250$
and $T=500$, which gives $66$ queries with probabilities between
$2\times10^{-11}$ and $10^{-6}$.
Forward simulation from $A$ never forms a droplet within $T$ steps,
whereas most relaxations from $B$ lose their droplet and are in $A$
at $T=500$ (Figure~\ref{fig:ac_relax}).

\begin{figure}[t]
\centering
\includegraphics[width=\columnwidth]{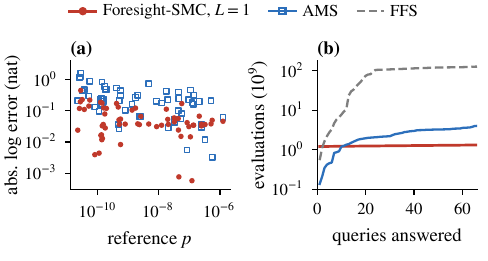}
\caption{Allen--Cahn field, one trained model for all
queries. (a)~Absolute log error of the $60$ point queries at equal
total cost. (b)~Cumulative evaluations of $U$ and $\nabla U$:
Foresight-SMC pays a one-off cost, whereas AMS and FFS, run to
$20\,\%$ RSE, pay again for every query.}
\label{fig:ac_queries}
\end{figure}

\paragraph{Accuracy and cost.}
We compare with adaptive multilevel splitting
\citep[AMS;][]{cerou2007adaptive,brehier2016unbiasedness} and forward
flux sampling \citep[FFS;][]{allen2005sampling}, which split
trajectories along a reaction coordinate, here the droplet size.
The cost of Foresight-SMC includes the one-off cost of generating
relaxations and training.
After this one-off cost, a query needs far fewer evaluations
than AMS run to $20\,\%$ relative standard error (RSE;
Figure~\ref{fig:ac_queries}b).
At equal total cost, the mean absolute log error
$E_{\log}=\operatorname{mean}\lvert\log\hat p-\log p(\nu,T)\rvert$ of
AMS is substantially larger than that of Foresight-SMC (Table~\ref{tab:ac},
Figure~\ref{fig:ac_queries}a).
FFS reaches a similar accuracy only at a much higher cost, because splitting starts over for every query.
In wall-clock time, Foresight-SMC takes about one hour in total on one
NVIDIA A100 GPU (Appendix~\ref{app:ac_cost}).

\paragraph{Effect of resampling.}
The results above use the learned proposal without resampling
($L=1$).
This proposal is close to the conditioned dynamics: its path KL is a
small fraction of that of the reference dynamics
(Table~\ref{tab:ac_seeds}).
The single-block weights therefore stay balanced, and resampling every
$50$ steps with the twist $e^{g_\theta}$ only adds twist error and
increases $E_{\log}$ (Table~\ref{tab:ac}), as
Section~\ref{sec:horizon} predicts.
Conversely, reference-proposal SMC, which keeps the twist but replaces
the learned proposal by the reference dynamics, returns zero estimates
on some point queries, so the accuracy comes from the learned
proposal.

\paragraph{Transition channels.}
A droplet can form at either end of the chain or in its interior.
Foresight-SMC recovers the share of $p(\nu,T)$ carried by each channel
in close agreement with the reference for all but one start (Table~\ref{tab:ac_channels},
Figure~\ref{fig:ac_generation}).

\begin{table}[t]
\caption{Allen--Cahn field. Cost: $10^9$ evaluations for all
$66$ queries, including the one-off cost of Foresight-SMC. $E_{\log}$:
over the $60$ point queries; Foresight-SMC gives mean $\pm$ SE over
three training seeds. In 2\,SE: point queries within two standard
errors of the reference. $\infty$: zero estimates on some queries.
Further baselines: Appendix~\ref{app:ac}.}
\label{tab:ac}
\centering
\small
\setlength{\tabcolsep}{3.5pt}
\begin{tabular}{@{}lrrr@{}}
\toprule
Method & Cost & $E_{\log}$ (nat) $\downarrow$ & In 2\,SE \\
\midrule
Foresight-SMC, $L=1$          & 1.32 & $0.070\pm0.002$ & 56--58 \\
Foresight-SMC, 50-step blocks & 1.31 & $0.119\pm0.027$ & 52--60 \\
Reference-proposal SMC        & 1.31 & $\infty$ & -- \\
\midrule
AMS, equal cost               & 1.29 & $0.296$ & 48 \\
AMS, to $20\,\%$ RSE          & 4.02 & $0.175$ & 51 \\
FFS, to $20\,\%$ RSE          & 127  & $0.059$ & 57 \\
Naive Monte Carlo             & 0.11 & $\infty$ & -- \\
\bottomrule
\end{tabular}
\end{table}

\begin{figure}[t]
\centering
\includegraphics[width=\columnwidth]{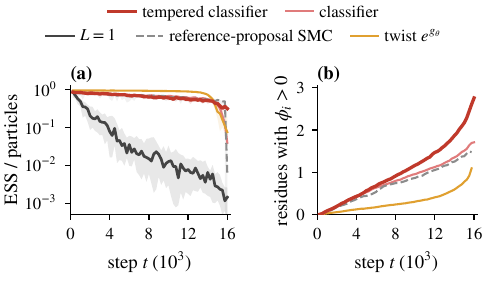}
\caption{Alanine tripeptide, eight broad queries.
(a)~Effective sample size of the weights along the horizon (median and
10--90\,\% band over queries), over the whole path for $L=1$ and per
block otherwise. (b)~Mean number of residues with $\phi_i>0$ among the
resampled particles. $E_{\log}$ of each variant: Table~\ref{tab:ala3}.}
\label{fig:ala3_resampling}
\end{figure}

\begin{figure}[t]
\centering
\includegraphics[width=\columnwidth]{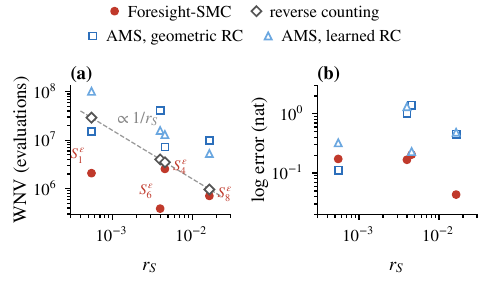}
\caption{Alanine tripeptide, four narrow start sets,
against the fraction $r_S$ of relaxations that end in each set.
(a)~Work-normalized variance; the cost of reverse counting grows as
$1/r_S$ (dashed). (b)~Absolute log error against the reverse-counting
reference. RC: reaction coordinate.}
\label{fig:ala3_narrow}
\end{figure}

\subsection{Alanine tripeptide}
\label{sec:exp_ala3}

Alanine tripeptide (Ace-Ala$_3$-Nme) is a small peptide whose backbone
conformation is described by the dihedral angles $(\phi_i,\psi_i)$ of
its three residues.
Each residue is mostly found with $\phi_i<0$. Conformations with
$\phi_i>0$ are rare because reaching them requires crossing a
free-energy barrier, so a transition in which all three residues flip
is a rare event.
The state $x\in\mathbb{T}^6$ collects the six angles. The potential $U$
is fitted to molecular dynamics data so that $\pi\propto e^{-U}$
reproduces the equilibrium distribution of the angles, and the
reference dynamics is MALA on $U$ (Appendix~\ref{app:ala3}).
The sets $A$ and $B$ contain the states with all three residues at
$\phi_i<0$ and at $\phi_i>0$, respectively, with
$\rho=\pi(B)/\pi(A)=1.3\times10^{-8}$, and the horizon is $T=16000$
steps.
Although $\rho$ is small, a large fraction of the relaxations are in
$A$ after $T$ steps.

The queries start from equilibrium restricted to subsets of $A$,
$\nu_S=\pi(\cdot\mid A\cap S)$.
Eight broad sets $S_1,\dots,S_8$, defined by the angles $\psi_i$,
partition $A$ and have $p$ of order $10^{-9}$.
Four narrow sets, small balls that each hold about $0.3\,\%$ of the
mass of $A$, act as localized starts that still admit an independent
reference; their centers serve as point starts.
Foresight-SMC resamples every $250$ steps ($L=64$) with the classifier
twist and averages $8$ independent runs of $512$ particles, called
islands, whose spread gives the standard error.

Because these start laws are restricted equilibrium laws and $\rho$ is
known here, Proposition~\ref{prop:path_reversal} with $A$ replaced by
$A\cap S$ gives a direct baseline, reverse counting:
\begin{equation}
\label{eq:reverse_count}
p(\nu_S,T)=\frac{\rho\,r_S}{f_S},
\qquad
r_S=P_{\mu_B}(Y_T\in A\cap S),
\end{equation}
where $f_S=\pi(S\mid A)$ and $r_S$ is estimated by the fraction of
relaxations that end in $A\cap S$.
The cost of reverse counting grows as $1/r_S$. It also requires
$\rho$, which is usually unknown, and does not apply to point starts.

\paragraph{Broad start sets.}
On this system the learned proposal is far from the conditioned
dynamics: its path KL remains a large fraction of that of the
reference dynamics (Appendix~\ref{app:ala3_results}).
Over $16000$ steps the weights without resampling therefore degenerate
(Figure~\ref{fig:ala3_resampling}a), and Foresight-SMC with $L=1$ has a
large error (Table~\ref{tab:ala3}).
This is the regime of Section~\ref{sec:horizon} in which resampling
reduces the variance.
Resampling restores the accuracy only when the learned proposal and the
classifier twist are combined: the proposal moves the particles toward
$B$, the twist keeps those that are likely to complete the transition,
and the particles enter the rare conformation steadily over the horizon
(Figure~\ref{fig:ala3_resampling}b).
With the reference proposal, particles rarely reach $B$ and most
islands return a zero estimate.
With the twist $e^{g_\theta}$, the error is markedly larger than with the
classifier twist, since the training loss determines $g_\theta$ only
through its gradient and its differences between neighboring states.
Tempering the classifier, with exponents set by pilot runs, lowers the
error further (Appendix~\ref{app:ala3_results}).
AMS splits trajectories along a reaction coordinate (RC), either a
geometric one built from the angles $\phi_i$ or a learned one built
from the classifier; both variants cost less but have a much larger
log error.
Reverse counting is the most accurate method on these sets, because
each broad set receives a large share of the returning relaxations.
The classifier plug-in of Section~\ref{sec:twist} also has a lower log
error, but it requires $\pi(B)$ and has no error bar
(Appendix~\ref{app:ala3_results}).

\paragraph{Narrow start sets.}
The narrow sets test localized starts, which few relaxations reach: the
four sets differ widely in $r_S$.
We compare the per-query work needed for a given precision through the
work-normalized variance $\mathrm{WNV}=C\cdot\mathrm{RSE}^2$, where $C$
is the cost per query and one-off costs are excluded.
Foresight-SMC has a lower average log error than both AMS variants and
a lower WNV on every set (Table~\ref{tab:ala3},
Figure~\ref{fig:ala3_narrow}).
The cost of Foresight-SMC does not depend on $r_S$, because it
simulates forward from the start set and uses the relaxations only for
training.
The cost of reverse counting, in contrast, grows as $1/r_S$, so it
needs considerably more work than Foresight-SMC on the two sets with
the smallest $r_S$ and comparable work on the other two.
The error of AMS varies strongly between sets and is large on some
of them (Table~\ref{tab:ala3_narrow}).
As on the broad sets, the variants without resampling, with the twist
$e^{g_\theta}$ or with the reference proposal are less accurate and
need more work, naive Monte Carlo finds no hit, and the classifier
plug-in is more accurate but provides no error bar.
The advantage of reverse counting on the broad sets thus disappears for
localized start sets, whereas the cost of Foresight-SMC stays the same.

\paragraph{Point starts.}
From the centers of the narrow sets, Foresight-SMC agrees closely with
a reference from an independent Foresight-SMC run with $64$ islands and
the untempered twist.
Reverse counting on shrinking balls around three of the centers agrees
with this reference (Appendix~\ref{app:ala3_results}).
Together, the two systems confirm the prediction of
Section~\ref{sec:horizon}: a single block suffices when the proposal is
accurate, and resampling with the learned twist is needed when the
weights degenerate over long horizons.

\begin{table}[t]
\caption{Alanine tripeptide. Cost: $10^7$ evaluations per broad
query, excluding one-off costs; on the narrow sets, costs are similar
except for learned-RC AMS (Appendix~\ref{app:ala3_estimators}).
$E_{\log}$: mean absolute log error (nat); $\pm$: SE over three
training seeds. WNV ratio: work-normalized variance relative to the
first row, range over the four narrow sets. The plug-in gives no error
bar; reverse counting is the narrow-set reference. $^\dagger$:
effective sample size below $12$, so the WNV is underestimated.
Per-set values: Tables~\ref{tab:ala3_narrow}
and~\ref{tab:ala3_broad}.}
\label{tab:ala3}
\centering
\footnotesize
\setlength{\tabcolsep}{1.7pt}
\begin{tabular}{@{}lcccc@{}}
\toprule
 & \multicolumn{2}{c}{Broad (8 sets)} & \multicolumn{2}{c}{Narrow (4 sets)}\\
\cmidrule(lr){2-3}\cmidrule(l){4-5}
Method & Cost & $E_{\log}\downarrow$ & $E_{\log}\downarrow$ & WNV ratio$\downarrow$ \\
\midrule
\multicolumn{5}{@{}l}{Foresight-SMC, twist:}\\
\quad tempered classifier & 6.6 & $0.13\pm0.03$ & $0.15$ & $1$ \\
\quad classifier          & 6.6 & $0.18\pm0.01$ & $0.16$ & 0.45--5.8 \\
\quad $e^{g_\theta}$      & 6.6 & $0.47$ & $0.34$ & 3.3--30 \\
Foresight-SMC, $L=1$      & 6.6 & $1.26$ & $1.03$ & 3.3--17$^\dagger$ \\
Reference-proposal SMC    & 6.6 & $1.03$ & $0.53$ & 9.5--75 \\
Classifier plug-in        & 0   & $0.07$ & $0.08$ & -- \\
\midrule
AMS, geometric RC         & 1.5 & $0.77$ & $0.74$ & 2.8--107 \\
AMS, learned RC           & 4.2 & $0.81$ & $0.59$ & 5.1--49 \\
Naive Monte Carlo         & 13.9 & $\infty$ & $\infty$ & $\infty$ \\
Reverse counting          & 6.6 & $0.03$ & ref. & 1.3--14 \\
\bottomrule
\end{tabular}
\end{table}

\section{Related Work}
\label{sec:related}

\paragraph{Rare-event estimators.}
Splitting methods \citep{allen2005sampling,cerou2007adaptive} and SMC
samplers for rare events \citep{cerou2012sequential} move particles
through intermediate levels of a reaction coordinate, and their
efficiency depends on that coordinate; importance sampling instead
needs a bias close to the conditioned dynamics, usually derived from
large-deviation theory \citep{dupuis2004importance}.
Foresight-SMC replaces the levels by a learned twist and the analytic
bias by a learned proposal, at the price of requiring equilibrium
samples of $\mu_B$ and $\pi$.

\paragraph{Learned proposals and twists.}
Stochastic control \citep{hartmann2012efficient,du2026rare}, the
cross-entropy method \citep{zhang2014applications} and learned SMC
\citep{heng2020controlled,lawson2022sixo} approximate the Doob
transform from forward simulations.
When the event is rare, these simulations contain few successes; on
alanine tripeptide, controlled SMC receives no training signal
(Appendix~\ref{app:learned}).
Neural committors \citep{khoo2019committor,strahan2023predicting},
which approximate the probability of reaching $B$ before $A$, estimate
a quantity close to our twist but return no probability estimate with
an error bar.
Learned path samplers \citep{holdijk2023stochastic,du2024doob} produce
transition paths without exact weights.
Foresight-SMC learns from the target side and places the learned
components in an unbiased estimator through exact likelihood ratios.

\paragraph{Time reversal.}
Time reversal under reversibility is classical in transition path
theory and kinetics \citep{e2006towards,hummer2004transition}.
Reverse-time SMC \citep{koskela2018inference} and learned diffusion
bridges \citep{heng2025simulating} approximate a backward process.
Under reversibility, no approximation is needed, because reversed
relaxations are exact samples of the conditioned path law; we use them
to train a forward twisted SMC estimator for any start law.

\section{Conclusion}
\label{sec:conclusion}

Foresight-SMC estimates rare endpoint probabilities of reversible
dynamics with twisted SMC whose proposal and twist are learned from
reversed relaxations.
Because one trained model serves all queries, on point queries and
localized start sets it needs less work per query for a given
precision than splitting, which starts over for every query.
When the learned proposal alone is accurate, as on the Allen--Cahn
field, a single block suffices; on alanine tripeptide, resampling with
the learned twist controls the weight degeneracy that accumulates over
long horizons.

The approach requires reversible dynamics, equilibrium samples of
$\mu_B$ and, for the classifier twist, equilibrium samples of $\pi$.
It targets the endpoint event $X_T\in B$, not a first-passage event.
Its rarity must also lie mainly in $\pi(B)/\pi(A)$: when $T$ is short
compared with the residence time in $B$, few relaxations return and
the training data are scarce.
For start laws other than $\mu_A$, including point starts, accuracy
relies on generalization.
Extensions to non-reversible dynamics and all-atom systems remain open.

\section*{Impact Statement}

This paper presents a method for estimating rare-event probabilities
in simulated stochastic systems, with applications in molecular
simulation. We do not foresee specific negative societal consequences
beyond those of the underlying simulations.

\bibliographystyle{icml2026}
\bibliography{references}

\clearpage
\appendix
\counterwithin{figure}{section}
\counterwithin{table}{section}
\onecolumn

\section{Path Reversal and Coverage}
\label{app:relaxation}

This appendix proves Proposition~\ref{prop:path_reversal} and the
coverage bound of Section~\ref{sec:relaxation}.

Throughout this section, $K$ is reversible with respect to $\pi$.
We write $\bar\mu_s=\mu_BK^s$ for the law of the relaxation at age $s$.

\subsection{Proof of Proposition~\ref{prop:path_reversal}}
\label{app:path_reversal}

By reversibility, for any $s\ge0$,
\begin{equation}
\label{eq:app_relaxation_density}
\bar\mu_s(dx)=\frac{h_s(x)}{\pi(B)}\,\pi(dx).
\end{equation}
Integrating over $A$ with $s=T$ gives
\begin{equation}
\label{eq:app_return_probability}
\pi(B)r_B=\int_A h_T(x)\,\pi(dx)=\pi(A)\,\mu_A(h_T).
\end{equation}
For a path $(y_0,\dots,y_T)$, detailed balance implies
\begin{equation}
\label{eq:app_path_identity}
\mu_B(dy_0)\prod_{t=0}^{T-1}K(y_t,dy_{t+1})
=\frac{\mathbf{1}_B(y_0)}{\pi(B)}\,\pi(dy_T)
\prod_{t=0}^{T-1}K(y_{t+1},dy_t).
\end{equation}
The right-hand side is the law of the reference chain started from $\pi$ at $y_T$ and run backward to $y_0$, restricted to $\{y_0\in B\}$ and divided by $\pi(B)$.
Conditioning on $Y_T\in A$ and using
\eqref{eq:app_return_probability}, we obtain
\begin{equation}
\operatorname{Law}(Y_T,\dots,Y_0\mid Y_T\in A)
=P_{\mu_A}(X_{0:T}\in\cdot\mid X_T\in B)
=P^\star_{\mu_A}.
\end{equation}
In particular, the initial marginal of the reversed paths is
\begin{equation}
P^\star_{\mu_A}(X_0\in dx)=\frac{h_T(x)}{\mu_A(h_T)}\,\mu_A(dx).
\end{equation}

\subsection{Reversal of records}
\label{app:records}

For an accepted MALA transition $x\to y$, detailed balance gives
\begin{equation}
\pi(x)\,q(x,y)\,\alpha(x,y)=\pi(y)\,q(y,x)\,\alpha(y,x),
\end{equation}
so the reversed transition $y\to x$, recorded as $\xi=(x,1)$, has the same weight, $\pi(y)R(y,(x,1))=\pi(x)R(x,(y,1))$.
A rejected transition at $x$ with proposal $z$ is reversed by
keeping $z$ and $a=0$; its weight $\pi(x)q(x,z)[1-\alpha(x,z)]$ is
unchanged.
For a fixed acceptance pattern, reversal permutes the coordinates
$(y_0,z_{0:T-1})$ and so preserves the reference measure. On the
torus, where records store increments, reversal negates the increments
of accepted steps and reverses their order; this map has unit Jacobian, so the argument is unchanged.
Equation~\eqref{eq:app_path_identity} therefore also holds for
recorded paths, and the reversed records follow the record law of
$P^\star_{\mu_A}$.

\subsection{Coverage}
\label{app:coverage}

Let $\nu^\star_t$ be the time-$t$ marginal of $P^\star_{\mu_A}$.
By the Markov property,
\begin{equation}
\nu^\star_t(dx)=\frac{(\mu_AK^t)(dx)\,h_{T-t}(x)}{\mu_A(h_T)}.
\end{equation}
Reversibility gives
\begin{equation}
(\mu_AK^t)(dx)=\frac{K^t\mathbf{1}_A(x)}{\pi(A)}\,\pi(dx).
\end{equation}
Combining these with \eqref{eq:app_relaxation_density} and
\eqref{eq:app_return_probability} yields
\begin{equation}
\nu^\star_t(dx)
=\frac{K^t\mathbf{1}_A(x)\,h_{T-t}(x)}{\pi(A)\,\mu_A(h_T)}\,\pi(dx)
=\frac{K^t\mathbf{1}_A(x)}{r_B}\,\bar\mu_{T-t}(dx).
\end{equation}
Since $0\le K^t\mathbf{1}_A\le1$,
\begin{equation}
\label{eq:app_coverage}
\frac{d\nu^\star_t}{d\bar\mu_{T-t}}\le\frac{1}{r_B}.
\end{equation}
Let $w=d\nu^\star_t/d\bar\mu_{T-t}$. Since $\mathbb{E}_{\bar\mu_{T-t}}[w]=1$, $\mathbb{E}_{\bar\mu_{T-t}}[w^2]\le\sup w\le1/r_B$. Relaxation states at age $T-t$, used as importance samples of $\nu^\star_t$, therefore have a population effective sample size fraction $1/\mathbb{E}_{\bar\mu_{T-t}}[w^2]$ of at least $r_B$.
The same bound holds for the path segment $X_{t:T}$. By reversibility, the backward kernel of the relaxation from age $s$ to $s-1$ is $K(x,dy)\,h_{s-1}(y)/h_s(x)$, the Doob kernel with $s$ steps remaining. The law of $X_{t:T}$ under $P^\star_{\mu_A}$ and that of the corresponding reversed relaxation segment therefore differ only through their marginals at time $t$.

\section{Controlled Kernel and Training}
\label{app:control}

This appendix derives the controlled kernel of
Section~\ref{sec:controlled_kernel} and the KL
identity~\eqref{eq:kl_identity}.

\subsection{Tilted MALA step}
\label{app:implementation}

Let $g=g_\theta(s,\cdot)$, where $s$ is the number of steps remaining after the step. Tilting one reference step by $e^{g(y)}$ and writing the tilted kernel in terms of the proposal $z$ gives
\begin{equation}
K(x,dy)\,e^{g(y)}
=\int q(x,z)\Big[\alpha(x,z)\,e^{g(z)}\,\delta_z(dy)
+\big(1-\alpha(x,z)\big)\,e^{g(x)}\,\delta_x(dy)\Big]dz.
\end{equation}
Given $z$, the step moves to $z$ with probability
$\alpha e^{g(z)}/\big(\alpha e^{g(z)}+(1-\alpha)e^{g(x)}\big)$,
which is \eqref{eq:controlled_acceptance}.
The proposal marginal is proportional to
$q(x,z)\big[\alpha e^{g(z)}+(1-\alpha)e^{g(x)}\big]$.
Ignoring rejections, we replace it by $q(x,z)e^{g(z)}$.
Expanding $g$ to first order around $x$,
\begin{align}
q(x,z)\,e^{g(z)}
&\propto\exp\!\Big(-\frac{\|z-x+\tau\nabla U(x)\|^2}{4\tau}
+\nabla g(x)\cdot z\Big)\nonumber\\
&\propto\mathcal{N}\big(z;\,x-\tau\nabla U(x)+2\tau\nabla g(x),\,2\tau I\big),
\end{align}
which is \eqref{eq:controlled_proposal}.
In our experiments the proposal is a two-component mixture,
\begin{equation}
\label{eq:app_mixture}
q^\theta_s(x,z)=\sum_{m=1}^{2}\lambda_{s,m}(x)\,
\mathcal{N}\big(z;\,x-\tau\nabla U(x)+2\tau b_{s,m}(x),\,2\tau I\big),
\end{equation}
where $b_{s,1}=\nabla_x g_\theta(s,x)$, $b_{s,2}=b_{s,1}+c_s(x)$,
$\lambda_{s,1}$ is the logistic function of a network output,
$\lambda_{s,2}=1-\lambda_{s,1}$, and $c_s$ is another network output.

For finite $g_\theta$, $q^\theta_s$ is positive everywhere, and
$\alpha^\theta_s\in(0,1)$ whenever $\alpha\in(0,1)$, with
$\alpha^\theta_s=1$ when $\alpha=1$.
The reference and controlled record laws are therefore mutually
absolutely continuous, and each term of $G_\theta$ is
\begin{equation}
\label{eq:app_record_ratio}
\log\frac{R^\theta_t(x,\xi)}{R(x,\xi)}
=\log\frac{q^\theta_s(x,z)}{q(x,z)}
+a\log\frac{\alpha^\theta_s(x,z)}{\alpha(x,z)}
+(1-a)\log\frac{1-\alpha^\theta_s(x,z)}{1-\alpha(x,z)}.
\end{equation}

\subsection{KL identity}
\label{app:kl}

Since $P^\theta_{\mu_A}$ and $P_{\mu_A}$ share the initial law
$\mu_A$,
\begin{equation}
\frac{dP^\theta_{\mu_A}}{dP_{\mu_A}}=e^{G_\theta}
\end{equation}
on recorded paths.
With \eqref{eq:conditioned_path_law},
\begin{align}
D_{\mathrm{KL}}\big(P^\star_{\mu_A}\,\|\,P^\theta_{\mu_A}\big)
&=\mathbb{E}_{P^\star_{\mu_A}}\Big[\log\frac{dP^\star_{\mu_A}}{dP_{\mu_A}}
-G_\theta\Big]\nonumber\\
&=-\log\mu_A(h_T)-\mathbb{E}_{P^\star_{\mu_A}}[G_\theta].
\end{align}
By Appendix~\ref{app:records}, the last expectation can be computed
from reversed relaxation paths, and
$\mu_A(h_T)=\pi(B)r_B/\pi(A)$ by \eqref{eq:app_return_probability}.
This gives \eqref{eq:kl_identity}.
Since the state path is a function of the recorded path, the data-processing inequality bounds the KL divergence between the state-path laws by this divergence.

\section{Twisted SMC}
\label{app:smc}

This appendix proves Theorem~\ref{thm:unbiased} and the scale
invariance of the twist used in Section~\ref{sec:twist}.

\subsection{Proof of Theorem~\ref{thm:unbiased}}
\label{app:unbiased}

Let $P_k(x,\cdot)$ and $Q_k(x,\cdot)$ denote the reference and
controlled record laws over block $k$, so that
\begin{equation}
\Lambda_k=\frac{dP_k}{dQ_k}=e^{-\Delta G_k},
\qquad
u_k=\Lambda_k\,\frac{\widehat H_k(X_{t_k})}{\widehat H_{k-1}(X_{t_{k-1}})}.
\end{equation}

\paragraph{Target identity.}
Since $\widehat H_0\equiv1$ and $\widehat H_L=\mathbf{1}_B$, the twist ratios telescope and
\begin{equation}
\prod_{k=1}^{L}u_k=e^{-G_\theta}\,\mathbf{1}_B(X_T).
\end{equation}
Consequently,
\begin{equation}
Z_L:=\mathbb{E}_{P^\theta_\nu}\Big[\prod_{k=1}^{L}u_k\Big]
=\mathbb{E}_{P_\nu}\big[\mathbf{1}_B(X_T)\big]=p(\nu,T).
\end{equation}

\paragraph{Particle unbiasedness.}
For a nonnegative function $f$ of the history $\omega_{1:k}$ up to
time $t_k$, where $\omega_k$ is the recorded path segment of block $k$,
define
\begin{equation}
\gamma_k(f)=\mathbb{E}_{P^\theta_\nu}\Big[f(\omega_{1:k})\prod_{k'=1}^{k}u_{k'}\Big],
\qquad
\widehat\gamma_k(f)=\frac{\widehat Z_{k-1}}{N}\sum_{i=1}^{N}u_k^{(i)}f(\omega^{(i)}_{1:k}),
\end{equation}
where $\widehat Z_k=\prod_{k'=1}^{k}N^{-1}\sum_iu_{k'}^{(i)}$ and
$\widehat Z_0=1$, so that $\widehat\gamma_L(1)=\hat p$.
Let
\begin{equation}
(\mathcal{M}_{k+1}f)(\omega_{1:k})
=\int Q_{k+1}(X_{t_k},d\omega_{k+1})\,u_{k+1}\,f(\omega_{1:k+1}).
\end{equation}
Then $\gamma_{k+1}(f)=\gamma_k(\mathcal{M}_{k+1}f)$.
Let $\mathcal{F}_k$ denote the particle histories up to $t_k$ and
$W_k^{(i)}=u_k^{(i)}/\sum_{i'}u_k^{(i')}$ the normalized weights.
Resampling with $\mathbb{E}[O_k^{(i)}\mid\mathcal{F}_k]=NW_k^{(i)}$,
followed by independent propagation, gives
\begin{equation}
\mathbb{E}\big[\widehat\gamma_{k+1}(f)\mid\mathcal{F}_k\big]
=\widehat Z_k\sum_{i=1}^{N}W_k^{(i)}(\mathcal{M}_{k+1}f)(\omega^{(i)}_{1:k})
=\widehat\gamma_k(\mathcal{M}_{k+1}f).
\end{equation}
Independent initialization from $\nu$ gives
$\mathbb{E}[\widehat\gamma_1(f)]=\gamma_1(f)$. Induction over $k$, applied to all nonnegative $f$ simultaneously, yields $\mathbb{E}[\widehat\gamma_k(f)]=\gamma_k(f)$ for all $k$, where Tonelli's theorem permits infinite values.
Taking $f\equiv1$ at $k=L$,
\begin{equation}
\mathbb{E}[\hat p]=\gamma_L(1)=Z_L=p(\nu,T).
\end{equation}
The argument uses only the conditional mean of the offspring counts,
not their independence, and therefore covers systematic resampling. Systematic resampling draws a single $V\sim\mathrm{Unif}(0,1)$ and assigns to particle
$i$ the points $(V+m)/N$, $m=0,\dots,N-1$, that fall in
$[\bar W_{i-1},\bar W_i)$, where $\bar W_i=\sum_{i'\le i}W_k^{(i')}$ and $\bar W_0=0$.
The point $(V+m)/N$ is uniform on $[m/N,(m+1)/N)$, so it falls in
$[\bar W_{i-1},\bar W_i)$ with probability
$N\,\lvert[\bar W_{i-1},\bar W_i)\cap[m/N,(m+1)/N)\rvert$; summing over $m$
gives $\mathbb{E}[O_k^{(i)}\mid\mathcal{F}_k]=N\,W_k^{(i)}$.

\subsection{Scale invariance of the twist}
\label{app:classifier}

If $\widehat H_k$ is replaced by $c_k\widehat H_k$ with constants $c_k>0$ for $0<k<L$, and $c_0=c_L=1$, each $u_k^{(i)}$ is multiplied by $c_k/c_{k-1}$. The normalized weights do not change, and
\begin{equation}
\hat p\;\longmapsto\;\hat p\prod_{k=1}^{L}\frac{c_k}{c_{k-1}}=\hat p.
\end{equation}

By \eqref{eq:app_relaxation_density}, relaxation states of age $s$
have density $h_s/\pi(B)$ with respect to $\pi$.
For a classifier trained on $n_1$ such states against $n_0$
equilibrium states, the Bayes-optimal logit is
\begin{equation}
\log\frac{n_1\,\bar\mu_s(x)}{n_0\,\pi(x)}
=\log h_s(x)-\log\pi(B)+\log\frac{n_1}{n_0}.
\end{equation}
The offset does not depend on $x$, so by the rescaling invariance
above the exponentiated logit at $s=T-t_k$ can be used as
$\widehat H_k$ \citep[cf.][]{lawson2022sixo}.
The offset is also independent of $s$ when $n_1/n_0$ is the same at every age, as in our experiments. This property is used only for interpolation between trained ages and for the AMS score built from the logit (Appendix~\ref{app:ala3_estimators}).

\section{Variance Analysis}
\label{app:variance}

This appendix proves the variance statements of
Section~\ref{sec:horizon} and examines the choice of the number of
blocks $L$.

Write $H_k=h_{T-t_k}$, and let $\nu^\star_k:=\nu^\star_{t_k}$ denote the time-$t_k$ marginal of $P^\star_\nu$.
Assume $H_{k-1}>0$ on the support of the reference marginal at
$t_{k-1}$.
The conditioned law of block $k$ started at $x$ is
\begin{equation}
P^\star_k(x,d\omega)=P_k(x,d\omega)\,\frac{H_k(X_{t_k})}{H_{k-1}(x)},
\end{equation}
with $P_k$ and $Q_k$ as in Appendix~\ref{app:unbiased}, and we write $\chi^2_k(x)=\chi^2\big(P^\star_k(x,\cdot)\,\|\,Q_k(x,\cdot)\big)$ and $d_k(x)=D_{\mathrm{KL}}\big(P^\star_k(x,\cdot)\,\|\,Q_k(x,\cdot)\big)$.

\begin{proposition}
\label{prop:app_variance}
(a) For $L=1$,
$N\operatorname{Var}(\hat p)/p^2=\chi^2(P^\star_\nu\|P^\theta_\nu)
\ge e^{D_{\mathrm{KL}}(P^\star_\nu\|P^\theta_\nu)}-1$.
(b) If at every interior boundary the ancestor indices are drawn
i.i.d.\ from $W_k$ given $\mathcal{F}_k$ and the moment conditions of
\citet{chopin2004central} hold, then $\sqrt N(\hat p/p-1)$ is asymptotically normal with
variance $\sigma^2=\sum_{k=1}^{L}v_k$, where
\begin{equation}
\label{eq:app_vk}
1+v_k
=\mathbb{E}_{\nu^\star_{k-1}}\Big[\frac{\widehat H_{k-1}}{H_{k-1}}\Big]\,
\mathbb{E}_{\nu^\star_{k-1}}\Big[\frac{H_{k-1}}{\widehat H_{k-1}}\big(1+\chi^2_k\big)\Big],
\end{equation}
and $\sigma^2\ge\sum_k(e^{\kappa_k}-1)\ge L(e^{\kappa/L}-1)$ with
$\kappa_k=\mathbb{E}_{\nu^\star_{k-1}}[d_k]$ and $\kappa=\sum_k\kappa_k$.
\end{proposition}

\paragraph{Proof of (a).}
With $L=1$, $\hat p=N^{-1}\sum_i p\,w^{(i)}$, where
\begin{equation}
w=\frac{\mathbf{1}_B(X_T)\,e^{-G_\theta}}{p}
=\frac{dP^\star_\nu}{dP^\theta_\nu}.
\end{equation}
The $w^{(i)}$ are i.i.d.\ with mean one, so
\begin{equation}
\frac{N\operatorname{Var}(\hat p)}{p^2}
=\mathbb{E}_{P^\theta_\nu}[w^2]-1
=\chi^2\big(P^\star_\nu\,\|\,P^\theta_\nu\big).
\end{equation}
By Jensen's inequality,
\begin{equation}
\mathbb{E}_{P^\theta_\nu}[w^2]=\mathbb{E}_{P^\star_\nu}[w]
\ge\exp\mathbb{E}_{P^\star_\nu}[\log w]
=e^{D_{\mathrm{KL}}(P^\star_\nu\|P^\theta_\nu)}.
\end{equation}
The population effective sample size fraction
$(\mathbb{E}w)^2/\mathbb{E}w^2=1/(1+\chi^2)$ is therefore at most
$e^{-D_{\mathrm{KL}}(P^\star_\nu\|P^\theta_\nu)}$.

\paragraph{Proof of (b).}
The standard central limit theorem for SMC estimates of normalizing
constants \citep{chopin2004central,delmoral2004feynman}, applied with
one resampling step per block, gives
$\sqrt N(\hat p/p-1)\Rightarrow\mathcal{N}(0,\sigma^2)$ with
\begin{equation}
\sigma^2=\sum_{k=1}^{L}v_k,
\qquad
1+v_k=\frac{\mathbb{E}[F_k^2]}{\mathbb{E}[F_k]^2},
\qquad
F_k=u_k\,\mathbb{E}\Big[\prod_{k'>k}u_{k'}\,\Big|\,X_{t_k}\Big],
\end{equation}
where the outer expectations are over a particle in block $k$ in the
limit $N\to\infty$.
Because the twists telescope and $\Lambda_{k'}$ turns expectations under
$Q_{k'}$ into expectations under $P_{k'}$,
\begin{equation}
\mathbb{E}\Big[\prod_{k'>k}u_{k'}\,\Big|\,X_{t_k}\Big]
=\frac{H_k(X_{t_k})}{\widehat H_k(X_{t_k})},
\qquad
F_k=\frac{\Lambda_k\,H_k(X_{t_k})}{\widehat H_{k-1}(X_{t_{k-1}})}.
\end{equation}
In the limit, block $k$ starts from $X_{t_{k-1}}$ with law
proportional to $(\widehat H_{k-1}/H_{k-1})\,\nu^\star_{k-1}$ and follows
$Q_k$.
Since $\Lambda_kH_k/H_{k-1}=dP^\star_k/dQ_k$,
\begin{equation}
\mathbb{E}_{Q_k(x,\cdot)}\big[\Lambda_kH_k\big]=H_{k-1}(x),
\qquad
\mathbb{E}_{Q_k(x,\cdot)}\big[(\Lambda_kH_k)^2\big]=H_{k-1}(x)^2\big(1+\chi^2_k(x)\big).
\end{equation}
Averaging over the start of the block gives \eqref{eq:app_vk}.

Two cases follow from \eqref{eq:app_vk}.
If the proposal is exact, $\chi^2_k=0$ and
$1+v_k=\mathbb{E}_{\nu^\star_{k-1}}[\widehat H_{k-1}/H_{k-1}]\,
\mathbb{E}_{\nu^\star_{k-1}}[H_{k-1}/\widehat H_{k-1}]$, which exceeds one
unless $\widehat H_{k-1}\propto H_{k-1}$.
If the twist is exact, that is, $\widehat H_{k-1}\propto H_{k-1}$, then
$v_k=\mathbb{E}_{\nu^\star_{k-1}}[\chi^2_k]$.

For a general positive twist, the Cauchy--Schwarz inequality applied
to \eqref{eq:app_vk}, $1+\chi^2_k\ge e^{d_k}$ as in part~(a), and
Jensen's inequality give
\begin{equation}
1+v_k
\ge\Big(\mathbb{E}_{\nu^\star_{k-1}}\big[\sqrt{1+\chi^2_k}\,\big]\Big)^2
\ge\Big(\mathbb{E}_{\nu^\star_{k-1}}\big[e^{d_k/2}\big]\Big)^2
\ge e^{\kappa_k},
\qquad
\kappa_k=\mathbb{E}_{\nu^\star_{k-1}}[d_k].
\end{equation}
Summing over blocks and using the convexity of $\exp$,
\begin{equation}
\sigma^2\ge\sum_{k=1}^{L}\big(e^{\kappa_k}-1\big)
\ge L\big(e^{\kappa/L}-1\big),
\qquad
\kappa=\sum_{k=1}^{L}\kappa_k.
\end{equation}
By the chain rule for the KL divergence along the Markov law
$P^\star_\nu$,
$\kappa=D_{\mathrm{KL}}(P^\star_\nu\|P^\theta_\nu)-D_{\mathrm{KL}}(\nu^\star\|\nu)$,
which is the path KL for a point start.

We do not verify the moment conditions of \citet{chopin2004central} for the mixture proposals. Part~(b) does not cover the systematic resampling used in our experiments, whose offspring counts are not conditionally independent \citep{douc2005comparison}; the resulting estimator is still unbiased by Theorem~\ref{thm:unbiased}.

\subsection{Choice of the number of blocks}
\label{app:lsweep}

\begin{figure}[h]
\centering
\includegraphics{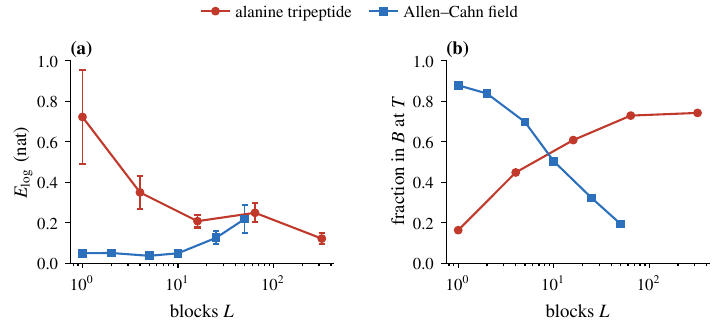}
\caption{Number of blocks $L$ at a fixed number of evaluations, mean
over four repetitions (bars: SE). (a)~Mean absolute log error:
alanine tripeptide, two broad queries, tempered classifier twist;
Allen--Cahn field, six point queries at $T=500$, twist $e^{g_\theta}$.
(b)~Fraction of particles in $B$ at $T$. Values differ from Tables~\ref{tab:ac} and~\ref{tab:ala3_broad}, which average over all queries.}
\label{fig:lsweep}
\end{figure}

In Figure~\ref{fig:lsweep}, $L$ varies at a fixed number of evaluations,
with four repetitions per setting. On alanine tripeptide, where the
path KL is about $11$ nat, $E_{\log}$ falls from $0.72$ at $L=1$ to
$0.1$--$0.25$ for $L\ge16$, and the variance of $\log\hat p$ across
islands falls from $3.6$ to about $0.3$. On the Allen--Cahn field, where
the path KL is $3.2$ nat and the twist is $e^{g_\theta}$, $E_{\log}$ on six point queries at $T=500$ is flat up to $L=10$ and grows for shorter blocks, mainly for start states
without a droplet, because fewer particles end in $B$. The two systems correspond to the two regimes of Section~\ref{sec:horizon}.

\section{Details of the Allen--Cahn Field}
\label{app:ac}

This appendix gives the details of Section~\ref{sec:ac};
Table~\ref{tab:hparams} lists the settings of both experiments.

\begin{table*}[t]
\centering
\caption{Settings for the Allen--Cahn field and alanine tripeptide.}
\label{tab:hparams}
\small
\begin{tabularx}{\linewidth}{@{}l>{\raggedright\arraybackslash}X>{\raggedright\arraybackslash}X@{}}
\toprule
 & Allen--Cahn field & Alanine tripeptide \\
\midrule
MALA step & $\tau=0.006$, $T\in\{250,500\}$ & $\tau=10^{-3}$ on unwrapped increments, $T=16000$ \\
Sampling $\mu_B$ & MALA restricted to $B$: 1024 chains, burn-in $6\times10^5$ steps, 128 draws per chain & exact: rejection sampling of $\phi$, then $\psi\sim\pi(\psi\mid\phi)$ \\
Relaxation pool & 131072 paths of 500 steps ($74\,\%$ in $A$ at $t=500$); train/validation/test $90/5/5$, split by chain & 32768 paths ($r_B=0.33$); held-out pool of 4096 for checkpoint selection \\
Records used & all ages, both horizons & 128 per returning path, ages uniform \\
$g_\theta$ & 1-D CNN, 4 layers $\times$ 64 channels, kernel 7; remaining steps enter through Fourier features & MLP, 4 layers $\times$ 256, SiLU; inputs $\cos$/$\sin$ of the angles, $U$, Fourier features of the remaining steps, $\mathbf{1}_B$ \\
Proposal & two-component mixture, Appendix~\ref{app:implementation} & same \\
Training & AdamW, lr $10^{-3}\to10^{-5}$ (cosine), batch 8192, 30000 steps, EMA 0.999; three seeds & same, 60000 steps; selected at 4000--6000 steps \\
Twist & $e^{g_\theta}$ & classifier: MLP $3\times128$, 21 ages, 131072 draws of $\pi$, 10000 Adam steps; tempering in Appendix~\ref{app:ala3} \\
Inference & $L=1$: batches of 4096 paths until $\ge50$ hits and RSE $\le20\,\%$; SMC: $8\times512$ particles, 50-step blocks & $8\times512$ particles, 250-step blocks; $L=1$: 4096 paths \\
AMS & 256 replicas, 26 removed per iteration; runs doubled from 32 to 512 until RSE $\le20\,\%$ & 256 replicas, $10\,\%$ removed; 5 runs per broad, 4 per narrow query \\
Reference & AMS with 1000 replicas, 1280 runs per query (median RSE $3.2\,\%$, max $21\,\%$) & reverse counting on independent pools of 32768 (broad) and 163840 (narrow) relaxations \\
One-off cost & $1.20\times10^9$ evaluations, almost all for sampling $\mu_B$ & $6.3\times10^8$ evaluations, plus $3.1\times10^9$ for the tempering pilot runs \\
\bottomrule
\end{tabularx}
\end{table*}

\subsection{System and queries}
\label{app:ac_system}

\begin{figure}[!htbp]
\centering
\includegraphics{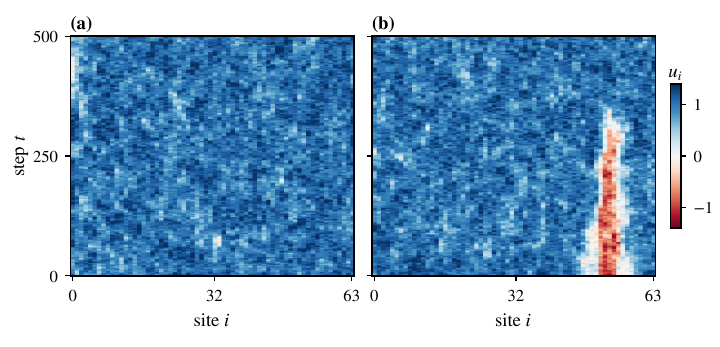}
\caption{Allen--Cahn field: reference dynamics as space--time plots of
the field $u_i$ (color; red: $u_i<0$, the unfavored phase) over the sites $i$ and the steps $t$.
(a)~From a point start in $A$ without a negative site, no droplet
forms within $T=500$. (b)~A relaxation from $\mu_B$: the droplet
shrinks and vanishes; read backwards, the relaxation is a training path.}
\label{fig:ac_relax}
\end{figure}

The state is a field $u=(u_0,\dots,u_{63})\in\mathbb{R}^{64}$ on a
chain with free ends, with energy
\begin{equation}
\label{eq:ac_energy}
\mathcal{E}(u)=\frac12\sum_{i=0}^{62}(u_{i+1}-u_i)^2
+\sum_{i=0}^{63}\Big[\frac{(1-u_i^2)^2}{4}-\eta\,u_i\Big].
\end{equation}
The first term penalizes interfaces, the second is a double well with
minima near $u_i=\pm1$, and the tilt $\eta=0.15$ favors $u_i=+1$.
The equilibrium law is $\pi\propto e^{-U}$ with $U=\mathcal{E}/\Theta$
at temperature $\Theta=0.15$.
The droplet size $\mathcal{D}(u)$ is the length of the longest run of consecutive sites with $u_i<0$ (negative sites). The sets are $A=\{\mathcal{D}\le3\}$ and $B=\{10\le\mathcal{D}\le16\}$.
The channel of a path that ends in $B$ is the position of this run at
time $T$: at the left end, at the right end, or in the interior. Figure~\ref{fig:ac_relax} shows the reference dynamics from a point start in $A$ and from $\mu_B$.

The $30$ point starts form five strata of six, defined by their negative sites: none, one, two adjacent, three adjacent in the interior, and
three adjacent at an end. Each point start is drawn from $\pi$ restricted to its sign pattern.
The start laws are $\pi$ restricted to $A_0=\{\mathcal{D}=0\}$,
$A_1=\{\mathcal{D}=1\}$ and $A_{23}=\{2\le\mathcal{D}\le3\}$.
Every start is queried at $T=250$ and $T=500$, which gives $60$ point queries and $66$ queries in total. For the three start laws, the estimates of Foresight-SMC lie within two standard errors of both an AMS and a reverse-counting reference.

\subsection{Baselines and reference values}
\label{app:ac_baselines}

All baselines are run by us with the same reference dynamics.

AMS is generalized fixed-horizon AMS \citep{brehier2016unbiasedness}
with the score $\min(\mathcal{D},10)$, set to $+\infty$ when
$X_T\in B$.
FFS uses the interfaces $\mathcal{D}=1,\dots,10$ and the same stopping rule as AMS (Table~\ref{tab:hparams}). Its cost includes its pilot runs.
Naive Monte Carlo spends the per-query budget of Foresight-SMC on
reference paths, and reference-proposal SMC uses the islands, blocks
and twist of Foresight-SMC with the reference kernel.
With $g_\theta$ as the reaction coordinate, AMS reaches $E_{\log}=0.155$, which is lower than with $\mathcal{D}$ and twice the error of Foresight-SMC, and it incurs the same one-off cost.
The reference probabilities come from AMS with the settings of
Table~\ref{tab:hparams}.
A point query counts as within two standard errors (``In 2\,SE'') if
the estimate and the reference differ by at most twice their standard
errors added in quadrature.
$E_{\log}$ also contains the error of the reference, and estimators
that stop at a target RSE carry a small stopping-rule bias.

\subsection{Additional results}
\label{app:ac_results}

\paragraph{Training seeds.}
The three seeds reach the same held-out path KL and similar accuracy with $L=1$ (Table~\ref{tab:ac_seeds}). With 50-step blocks, seeds 2 and 3 lose accuracy, as expected from Section~\ref{sec:horizon}.

\begin{table}[!htbp]
\centering
\caption{Allen--Cahn field: training seeds. KL is the held-out path KL
for the start law $\mu_A$; for comparison, $-\log p$ is $22.3$ nat at
$T=500$ and $23.3$ nat at $T=250$. $E_{\log}$ over the 60 point
queries, with the number of queries within two standard errors in parentheses.
Seed 1 has the best validation objective.}
\label{tab:ac_seeds}
\small
\begin{tabular}{@{}lcccc@{}}
\toprule
Seed & KL, $T=500$ & KL, $T=250$ & $E_{\log}$, $L=1$ & $E_{\log}$, 50-step blocks \\
\midrule
1 & $3.24\pm0.07$ & $5.67\pm0.17$ & $0.073$ (58) & $0.065$ (60) \\
2 & $3.27\pm0.07$ & $5.68\pm0.18$ & $0.067$ (56) & $0.146$ (55) \\
3 & $3.26\pm0.08$ & $5.67\pm0.17$ & $0.071$ (56) & $0.146$ (52) \\
\bottomrule
\end{tabular}
\end{table}

\begin{figure*}[!htbp]
\centering
\includegraphics[width=\linewidth]{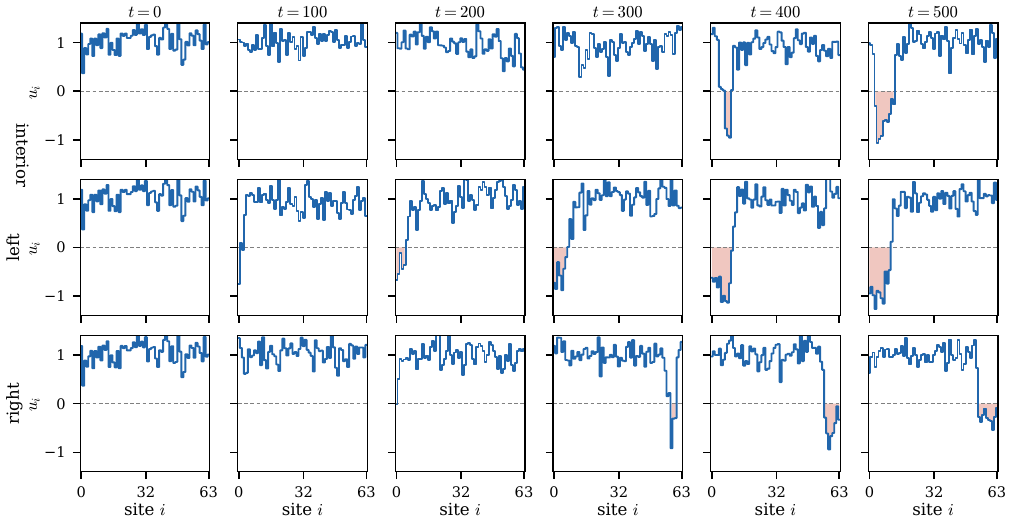}
\caption{Allen--Cahn field: paths generated by Foresight-SMC with
$L=1$ from the droplet-free point start, one per channel (rows:
interior, left end, right end), each the path of median importance
weight in its channel. Columns show the field $u_i$ at
$t=0,100,\dots,500$; the dashed line is $u=0$ and the shaded area
marks negative sites.}
\label{fig:ac_generation}
\end{figure*}

\paragraph{Transition channels.}
Point starts that already contain a negative run commit to the
corresponding channel, and Foresight-SMC reproduces the channel shares of the reference (Table~\ref{tab:ac_channels}). The largest total variation distance, $0.11$, occurs for the droplet-free start at $T=250$.
Figure~\ref{fig:ac_generation} shows one generated path per channel.

\begin{table}[!htbp]
\centering
\caption{Allen--Cahn field: channel shares (left end / right end /
interior) against the reference, and their total variation (TV).
Starts: $\mu_A$ (reference: reverse counting) and one point start of
each stratum, labeled by its number of negative sites (reference:
AMS).}
\label{tab:ac_channels}
\small
\begin{tabular}{@{}llccc@{}}
\toprule
$T$ & Start & Foresight-SMC & Reference & TV \\
\midrule
500 & $\mu_A$          & .397 / .436 / .167 & .394 / .381 / .225 & 0.058 \\
    & 0                & .423 / .285 / .292 & .396 / .316 / .288 & 0.031 \\
    & 1                & .347 / .259 / .394 & .408 / .228 / .364 & 0.060 \\
    & 2                & .009 / .005 / .986 & .010 / .010 / .981 & 0.005 \\
    & 3, interior      & .001 / .002 / .998 & .000 / .003 / .997 & 0.001 \\
    & 3, left end      & .990 / .002 / .008 & .991 / .000 / .009 & 0.001 \\
    & 3, right end     & .002 / .996 / .003 & .000 / .991 / .009 & 0.006 \\
\midrule
250 & $\mu_A$          & .333 / .363 / .304 & .327 / .313 / .359 & 0.056 \\
    & 0                & .374 / .159 / .467 & .421 / .218 / .361 & 0.106 \\
    & 1                & .061 / .048 / .891 & .030 / .026 / .944 & 0.053 \\
    & 2                & .000 / .000 / 1.000 & .000 / .000 / 1.000 & 0.000 \\
    & 3, interior      & .000 / .000 / 1.000 & .000 / .000 / 1.000 & 0.000 \\
    & 3, left end      & .994 / .000 / .006 & .991 / .001 / .008 & 0.003 \\
    & 3, right end     & .000 / .998 / .002 & .000 / .995 / .005 & 0.003 \\
\bottomrule
\end{tabular}
\end{table}

\subsection{Computational cost}
\label{app:ac_cost}

Foresight-SMC runs on one NVIDIA A100-PCIE-40GB GPU, and the splitting baselines run on single cores of a Huawei Kunpeng-920 (ARM) CPU.
Training takes $0.84$ GPU hours per seed, and answering all $66$ queries takes $0.16$ GPU hours. At $20\,\%$ RSE, AMS and FFS take $1.1$ and $36$ CPU core hours, respectively.

\section{Details of the Alanine Tripeptide Model}
\label{app:ala3}

This appendix gives the details of Section~\ref{sec:exp_ala3}; the
settings are listed in Table~\ref{tab:hparams}.

\subsection{System and queries}
\label{app:ala3_system}

\paragraph{Potential and dynamics.}
The state is $x=(\phi_1,\psi_1,\phi_2,\psi_2,\phi_3,\psi_3)\in\mathbb{T}^6$,
the backbone dihedral angles of Ace-Ala$_3$-Nme.
The potential $U$ is fitted to Langevin dynamics simulations of the molecule at
$300$\,K (Amber ff96, OBC implicit solvent
\citep{onufriev2004exploring}).
Since the rare conformation ($\phi_i>0$) is seldom visited, these simulations are
run with a torsional bias on $\phi$, which is removed by reweighting
\citep{shirts2008statistically}.
We write $\pi(x)=\pi(\phi)\,\pi(\psi\mid\phi)$: the marginal of $\phi$
is a smoothed histogram interpolated by a periodic spline, and
$\pi(\psi\mid\phi)$ is an autoregressive mixture of von Mises
densities fitted by maximum likelihood, so that $U=-\log\pi$ is smooth
and periodic.
The reference kernel is MALA on the unwrapped increment
$\delta=z-x$, after which the angles are wrapped to $\mathbb{T}^6$. Since the wrapped state does not determine the increment, the records store $(\delta,a)$. The network $g_\theta$ receives the angles through their sines and cosines.

\begin{figure*}[!htbp]
\centering
\includegraphics[width=\linewidth]{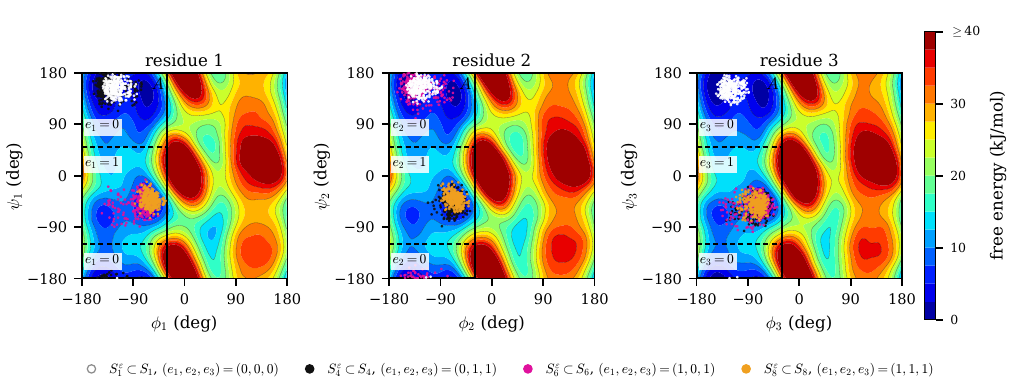}
\caption{Alanine tripeptide: start sets on the $(\phi_i,\psi_i)$
plane of each residue, over the marginal free energy (saturated above
$40$\,kJ\,mol$^{-1}$). The box is the range of $\phi_i$ in $A$; dashed
lines mark the band $\psi_i\in[-120^\circ,50^\circ)$, where $e_i=1$.
Each broad set $S_j$ fixes $(e_1,e_2,e_3)$. Points: 300 start states
of each narrow set $S^\varepsilon_j$.}
\label{fig:ala3_sets}
\end{figure*}

\paragraph{Sets and queries.}
The sets $A$ and $B$ are
\begin{equation}
A=\{x:\ \phi_i\in[-180^\circ,-30^\circ],\ i=1,2,3\},
\qquad
B=\{x:\ \phi_i\in[30^\circ,110^\circ],\ i=1,2,3\}.
\end{equation}
$B$ is metastable: of $128$ reference paths started from $\mu_B$,
$72\,\%$ are still in $B$ after $1000$ steps.
With $e_i(x)=1$ if $\psi_i\in[-120^\circ,50^\circ)$ and $e_i(x)=0$
otherwise, the broad sets are
\begin{equation}
S_j=\{x:\ 4e_1(x)+2e_2(x)+e_3(x)=j-1\},\qquad j=1,\dots,8.
\end{equation}
For $j\in\{1,4,6,8\}$, the narrow set $S^\varepsilon_j$ is the
geodesic ball in $A\cap S_j$ centered at a state $\bar x_j$, with $\pi(S^\varepsilon_j)\approx0.003\,\pi(A)$. The centers $\bar x_j$ are the point starts (Figure~\ref{fig:ala3_sets}).

\subsection{Training, baselines and reference values}
\label{app:ala3_estimators}

As on the Allen--Cahn field, all baselines are run by us with the same
reference dynamics, and one set of settings (Table~\ref{tab:hparams})
serves all queries.

\paragraph{Training data and twist.}
Draws of $\mu_B$ and of $\pi$ are exact, by rejection sampling of
$\phi$ followed by $\psi\sim\pi(\psi\mid\phi)$.
The classifier is a single network with input $(s,x)$, trained with
balanced classes to separate states of age $s=0,800,\dots,16000$ of
all training relaxations from draws of $\pi$. Between these ages, the exponentiated logits are interpolated linearly.
The tempered twist uses $\beta_k=\beta(T-t_k)$ with $\beta(s)$
piecewise linear in $\log s$ through $(s,\beta)=(250,1.86)$,
$(1000,1.45)$ and $(4000,1.19)$, and $\beta(s)=1$ for $s\ge12000$.
These values were fixed from six pilot runs with start states
disjoint from the reported ones. Since $\beta$ is fixed before the reported runs, Theorem~\ref{thm:unbiased} applies conditionally on $\beta$.

\paragraph{Baselines.}
Reference-proposal SMC uses the reference kernel with the twist and
blocks of Foresight-SMC.
AMS is fixed-horizon generalized AMS \citep{brehier2016unbiasedness}.
Its score is $\zeta(T,x)=\mathbf{1}_B(x)$ at the final time and, for
$t<T$,
\begin{equation}
\zeta(t,x)=0.998\exp\Big(-\sum_{i=1}^{3}\operatorname{dist}\big(\phi_i,[30^\circ,110^\circ]\big)-\lambda_0\big(1-t/T\big)\Big)
\end{equation}
for the geometric reaction coordinate, where $\operatorname{dist}$ is the angular distance to the interval, with $\lambda_0\in\{0,1,3\}$
selected by pilot runs, or $\zeta(t,x)=0.998/\big(1+e^{-\ell_{T-t}(x)}\big)$
for the learned one.

\paragraph{Reference values and costs.}
The ratio $\rho=1.32\times10^{-8}$ is computed by quadrature over
$\phi$.
The references are reverse counting on independent pools of
relaxations (Table~\ref{tab:hparams}), with RSE $2$--$5\,\%$ on the
broad and $2$--$11\,\%$ on the narrow sets. The broad reference shares $\rho$ and $f_S$ with the reverse-counting baseline, so the error of this baseline in Table~\ref{tab:ala3_broad} reflects counting noise only.
The one-off costs, excluded from the tables, are $5.2\times10^8$
evaluations for the relaxation pool, $0.7\times10^8$ for the held-out
pool and $0.4\times10^8$ for the classifier, which learned-RC AMS also requires. The tempering pilot runs cost $3.1\times10^9$ evaluations.
Table~\ref{tab:ala3_narrow} lists, for the four narrow sets, the log errors, the RSE of Foresight-SMC and the WNV of each baseline relative to Foresight-SMC. For the narrow sets, the RSE is computed across the eight islands for
Foresight-SMC, across four runs for AMS, and binomially for reverse
counting with $4096$ relaxations.
With eight islands, the WNV of Foresight-SMC is uncertain by a factor
of about two. With $32$ islands, bootstrap $95\,\%$ intervals of the
WNV ratio of reverse counting to Foresight-SMC exclude one on
$S^\varepsilon_1$ and $S^\varepsilon_6$.
If one relaxation pool served all start sets at once, reverse counting
would need less work than Foresight-SMC on $S^\varepsilon_4$ and
$S^\varepsilon_8$.

\begin{table}[!htbp]
\caption{Alanine tripeptide: eight broad queries with
diagnostics. Cost: $10^8$ evaluations for all eight queries, excluding
one-off costs (Appendix~\ref{app:ala3_estimators}). Diagnostic:
fraction of particles in $B$ at $T$ (SMC), effective sample size out
of $4096$ ($L=1$), islands with a zero estimate out of $64$, or RSE of
a single run. $\pm$: SE over three training seeds.}
\label{tab:ala3_broad}
\centering
\small
\setlength{\tabcolsep}{3.5pt}
\begin{tabular}{@{}lrrl@{}}
\toprule
Method & Cost & $E_{\log}$ (nat) $\downarrow$ & Diagnostic \\
\midrule
\multicolumn{4}{@{}l}{Foresight-SMC, twist:}\\
\quad tempered classifier & 5.24 & $0.13\pm0.03$ & 77\,\% in $B$ \\
\quad classifier          & 5.24 & $0.18\pm0.01$ & 18\,\% in $B$ \\
\quad $e^{g_\theta}$      & 5.24 & $0.47$ & 16\,\% in $B$ \\
Foresight-SMC, $L=1$      & 5.24 & $1.26$ & ESS 4--15 \\
Reference-proposal SMC    & 5.24 & $1.03$ & 46/64 zero \\
\midrule
AMS, geometric RC        & 1.20 & $0.77$ & RSE 30--120\,\% \\
AMS, learned RC          & 3.35 & $0.81$ & RSE 40--140\,\% \\
Naive Monte Carlo        & 11.1 & $\infty$ & no hit \\
Reverse counting         & 5.24 & $0.03$ & -- \\
\bottomrule
\end{tabular}
\end{table}

\begin{table}[!htbp]
\caption{Alanine tripeptide: four narrow start sets, ordered by
increasing $r_S$, at about $6.6\times10^7$ evaluations per query
(learned-RC AMS: two to three times more). WNV ratio: relative to
Foresight-SMC with the tempered classifier twist, the default variant;
above one means more work for the same precision. Reverse counting
provides the reference. Naive Monte Carlo finds no hit. $^\dagger$:
effective sample size below $12$, so the WNV is underestimated.}
\label{tab:ala3_narrow}
\centering
\small
\setlength{\tabcolsep}{4pt}
\begin{tabular}{@{}lcccc@{}}
\toprule
 & $S^\varepsilon_1$ & $S^\varepsilon_6$ & $S^\varepsilon_4$ & $S^\varepsilon_8$ \\
$10^3\,r_S$ & $0.55$ & $4.0$ & $4.6$ & $16$ \\
\midrule
\multicolumn{5}{@{}l}{Log error $\lvert\log\hat p-\log p\rvert$ (nat)}\\
Foresight-SMC    & 0.17 & 0.17 & 0.21 & 0.04 \\
Foresight-SMC, untempered classifier & 0.02 & 0.31 & 0.12 & 0.17 \\
Foresight-SMC, twist $e^{g_\theta}$ & 0.38 & 0.43 & 0.52 & 0.03 \\
Foresight-SMC, $L=1$ & 2.66 & 0.22 & 1.14 & 0.09 \\
Reference-proposal SMC & 0.44 & 0.77 & 0.77 & 0.13 \\
AMS, geometric RC & 0.11 & 1.01 & 1.40 & 0.45 \\
AMS, learned RC  & 0.32 & 1.32 & 0.23 & 0.49 \\
\midrule
RSE of Foresight-SMC (\%) & 17.8 & 7.7 & 19.7 & 10.4 \\
\midrule
\multicolumn{5}{@{}l}{WNV relative to Foresight-SMC}\\
AMS, geometric RC & 7.3  & 107  & 2.8  & 13.7 \\
AMS, learned RC  & 48.7 & 40.9 & 5.1  & 7.5 \\
Reverse counting & 14.0 & 10.4 & 1.4 & 1.3 \\
Foresight-SMC, untempered classifier & 1.1 & 5.8 & 0.45 & 0.58 \\
Foresight-SMC, twist $e^{g_\theta}$ & 10.8 & 29.6 & 3.3 & 6.6 \\
Foresight-SMC, $L=1^\dagger$ & 3.3 & 17.3 & 6.2 & 8.1 \\
Reference-proposal SMC & 11.9 & 74.6 & 9.5 & 69.8 \\
\bottomrule
\end{tabular}
\end{table}

\subsection{Additional results}
\label{app:ala3_results}

\paragraph{Path KL.}
The held-out path KL is $11.0\pm1.4$, $10.7\pm1.7$ and $11.8\pm1.3$ nat
for the three training seeds, against $-\log(\rho\,r_B)=19.2$ nat for
the reference dynamics.

\paragraph{Effect of tempering.}
The tempering parameters were fixed by pilot runs on the broad start laws. On fresh start states, the tempered and untempered twists give $E_{\log}$ of $0.10$ and $0.09$, respectively, on the broad queries. With tempering, the variance of $\log\hat p$ across islands is $0.43$ times that without. On the narrow sets, the two twists give comparable errors.

\paragraph{Point starts.}
For the point starts $\bar x_j$, the reference is Foresight-SMC with
the untempered twist and $64$ islands, which is unbiased by
Theorem~\ref{thm:unbiased}. Reverse counting on shrinking balls around $\bar x_4$, $\bar x_6$ and $\bar x_8$, which uses no learned component, agrees with this reference within one to two standard errors. Runs with $8$ islands have mean log errors of $0.12$ with the tempered and $0.15$ with the untempered twist (Table~\ref{tab:ala3_points}).

\begin{table}[!htbp]
\centering
\caption{Alanine tripeptide, point starts at the narrow-set centers:
reference $\log p$ (RSE) from an independent Foresight-SMC run (untempered twist, $64$ islands) and absolute log errors of Foresight-SMC ($8$ islands) with the two classifier twists and of the classifier
plug-in.}
\label{tab:ala3_points}
\small
\setlength{\tabcolsep}{3.5pt}
\begin{tabular}{@{}lcccc@{}}
\toprule
Start & Reference & Tempered & Untempered & Plug-in \\
\midrule
$\bar x_1$ & $-20.02$ ($6.0\,\%$) & 0.14 & 0.11 & 0.03 \\
$\bar x_4$ & $-17.72$ ($5.6\,\%$) & 0.05 & 0.16 & 0.07 \\
$\bar x_6$ & $-18.03$ ($6.6\,\%$) & 0.11 & 0.23 & 0.14 \\
$\bar x_8$ & $-16.42$ ($5.3\,\%$) & 0.18 & 0.11 & 0.09 \\
\bottomrule
\end{tabular}
\end{table}

\paragraph{Classifier plug-in.}
The plug-in estimate $\log\nu(e^{\ell_T})+\log\pi(B)$ has mean log
error $0.07$--$0.08$ on all three query types, below that of
Foresight-SMC.
These queries favor the plug-in: at $T=16000$ the logit varies little within a
start set, and the start laws are restrictions of $\pi$, which supplies the negative class of the classifier.
The plug-in requires $\pi(B)$ and has the limitations stated in Section~\ref{sec:twist}.

\begin{figure*}[!htbp]
\centering
\includegraphics[width=0.85\linewidth]{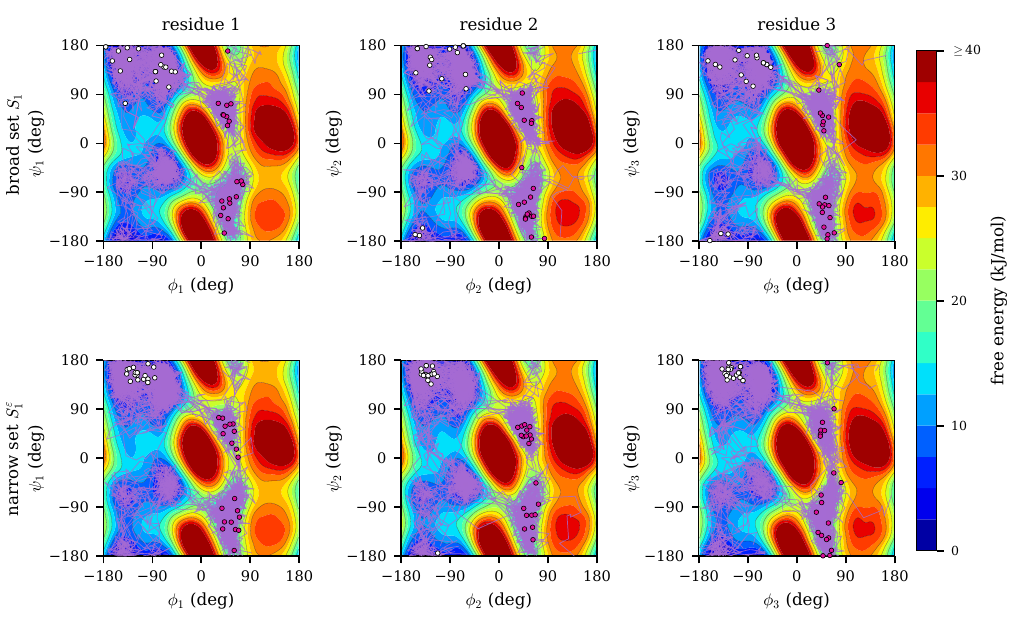}
\caption{Alanine tripeptide: paths of Foresight-SMC over
$T=16000$ steps from the broad set $S_1$ (top) and the narrow set
$S^\varepsilon_1$ (bottom), on the $(\phi_i,\psi_i)$ plane of each
residue (columns) over the free energy of Figure~\ref{fig:ala3_sets}.
Each row shows 20 paths from a start state (white) to an end state in
$B$ (magenta), chosen so that the end states follow the full sample.}
\label{fig:ala3_paths}
\end{figure*}

\begin{figure*}[!htbp]
\centering
\includegraphics[width=\linewidth]{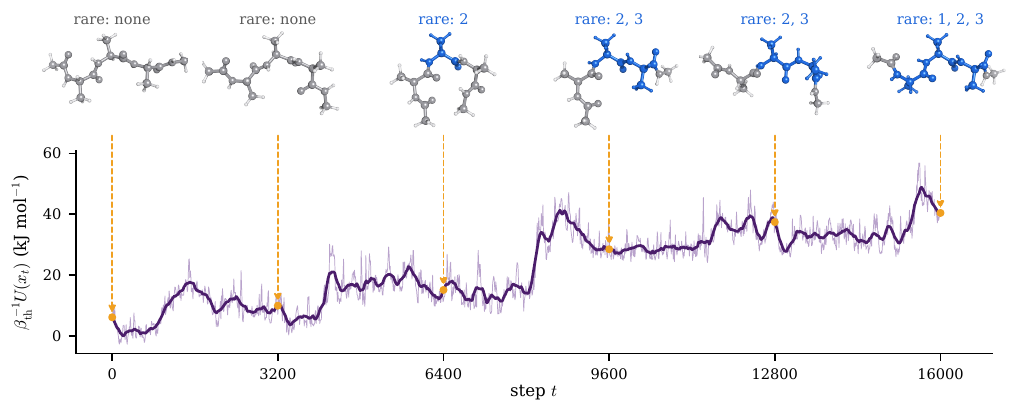}
\caption{Alanine tripeptide: potential energy
$\beta_{\mathrm{th}}^{-1}U(x_t)$, with
$\beta_{\mathrm{th}}^{-1}=k_{\mathrm{B}}\times300$\,K, shifted to a
minimum of zero, along one path from the top row of
Figure~\ref{fig:ala3_paths} (light: every tenth step; dark: running
mean over 201 steps). Above: all-atom structures at the marked steps,
with residues in the rare conformation in blue. The residues flip one
at a time, and each flip raises the potential. None of 32 reference
paths from the same start state flips a single residue within $T$.}
\label{fig:ala3_snapshots}
\end{figure*}

\paragraph{End states and generated paths.}
Proposition~\ref{prop:path_reversal} predicts the end-state law of conditioned paths from relaxations (Table~\ref{tab:ala3_ends}). Relative to $\mu_B$, this law favors the upper basin of $B$ for $S_1$ and the lower basin for $S_8$. For every residue and training seed, the fraction of end states of Foresight-SMC in the upper basin exceeds that under $\mu_B$ for $S_1$ and falls below it for $S_8$.
Figures~\ref{fig:ala3_paths} and~\ref{fig:ala3_snapshots} show
generated paths.
The free-energy backgrounds of Figures~\ref{fig:ala3_sets} and~\ref{fig:ala3_paths} are computed for the all-atom molecule by metadynamics \citep{barducci2008well}. The structures in Figure~\ref{fig:ala3_snapshots} are reconstructed from the six angles by restrained energy minimization.

\begin{table}[!htbp]
\centering
\caption{Alanine tripeptide: fraction of end states in the upper
basin of $B$ ($\psi_i\in[-30^\circ,150^\circ)$) for residues $1,2,3$.
Prediction: initial states $Y_0\sim\mu_B$ of relaxations, weighted by $\mathbf{1}_{A\cap S}(Y_T)$. Foresight-SMC: range over
three training seeds.}
\label{tab:ala3_ends}
\small
\begin{tabular}{@{}lccc@{}}
\toprule
 & Residue 1 & Residue 2 & Residue 3 \\
\midrule
$\mu_B$ & 0.26 & 0.25 & 0.23 \\
Prediction, $S_1$ & 0.40 & 0.43 & 0.38 \\
Foresight-SMC, $S_1$ & 0.40--0.53 & 0.32--0.54 & 0.38--0.53 \\
Prediction, $S_8$ & 0.09 & 0.10 & 0.09 \\
Foresight-SMC, $S_8$ & 0.10--0.13 & 0.08--0.12 & 0.09--0.16 \\
\bottomrule
\end{tabular}
\end{table}

\subsection{Computational cost}
\label{app:ala3_cost}

A controlled step, which also evaluates $g_\theta$, costs
$6.6$\,\textmu s per particle against $4.8$\,\textmu s for a reference step.
All computations for this system, including AMS, run on one NVIDIA A100-PCIE-40GB GPU. The relaxation pool takes $0.8$ GPU hours and training $0.2$ to $0.35$
GPU hours per seed. A narrow query takes $0.15$ to $0.35$ GPU hours with Foresight-SMC and $0.12$ to $0.48$ GPU hours with AMS, so Foresight-SMC is also cheaper than AMS per unit of variance in GPU time (Table~\ref{tab:ala3_narrow}).

\section{Comparison with Forward-Trained Samplers}
\label{app:learned}

We compare samplers whose proposals are learned without target-side data.
On the Allen--Cahn field, the cross-entropy sampler uses the network of Foresight-SMC but fits it to forward rollouts, with intermediate levels in the droplet size $\mathcal{D}$. With the one-off budget of Foresight-SMC, it returns a zero estimate on all $60$ point queries.
On alanine tripeptide, we adapt the diffusion path sampler TPS-DPS \citep{seong2025transition}, which learns a sampler for fixed end states, and trust-region transport \citep{blessing2025trust}, a path-space stochastic control method. Both run inside a MALA-type propose--accept step that records each transition, so that their likelihood ratios are exact, as for Foresight-SMC. Their proposals either rarely reach $B$ or produce degenerate weights, and their log errors are far larger than that of Foresight-SMC (Table~\ref{tab:learned}).

\paragraph{Controlled SMC.}
Controlled SMC \citep{heng2020controlled} learns the proposal by
approximate dynamic programming on its own particles.
To isolate the source of training data, we give it the kernel family of
Foresight-SMC with a single mixture component, the same twist form, blocks and resampling, and one policy shared by all start states. Only $g$ is learned, by regressing $\exp g$ on the block-wise target
$\exp(-\Delta G_{k+1})\,\widehat H_{k+1}$ backward from $\widehat H_L=\mathbf{1}_B$,
using all particles of the previous iteration.
On alanine tripeptide no forward particle reaches $B$, at the budget of
Foresight-SMC or at four times that budget, so every regression target
is zero and the method reduces to SMC with the reference kernel.

\begin{table}[h]
\centering
\caption{Alanine tripeptide: learned samplers without target-side
data. Hits are the fraction of paths that end in $B$; $\infty$: zero estimate.}
\label{tab:learned}
\small
\begin{tabular}{@{}lll@{}}
\toprule
Method & Hits in $B$ & $E_{\log}$ (nat) \\
\midrule
TPS-DPS                      & 4.4--6.8\,\% & 17.7 \\
Trust-region transport       & 0 of 32768  & $\infty$ \\
Foresight-SMC, $L=1$         & 14.5--29.4\,\% & 1.26 \\
Foresight-SMC, 250-step blocks, tempered classifier & 77\,\% & 0.13 \\
\bottomrule
\end{tabular}
\end{table}

\end{document}